\documentclass{article} 
\usepackage{arxiv,times}
\iclrfinalcopy

\usepackage{booktabs}       
\usepackage{xcolor}         
\usepackage{multirow}
\usepackage{enumitem}
\usepackage{extarrows}
\usepackage{algorithm}
\usepackage{algpseudocode}
\usepackage{graphicx}
\usepackage{subfigure}
\usepackage{caption}
\usepackage{wrapfig}
\usepackage{nicefrac}
\usepackage{makecell}
\usepackage{subcaption}
\usepackage{mwe}
\usepackage[most]{tcolorbox} 

\newtcolorbox{mybox}[1]{
  colback=bg,
  colframe=blue!75!black,
  fonttitle=\bfseries,
  title=#1
}

\newtcolorbox{generatedbox}[1]{
  colback=white,
  colframe=red!75!black,
  fonttitle=\bfseries,
  title=#1
}

\colorlet{bg}{white}

\usepackage{amsmath,amsfonts,bm}

\def\eqref#1{equation~\ref{#1}}

\def\1{\bm{1}}

\DeclareMathAlphabet{\mathsfit}{\encodingdefault}{\sfdefault}{m}{sl}
\SetMathAlphabet{\mathsfit}{bold}{\encodingdefault}{\sfdefault}{bx}{n}
\newcommand{\tens}[1]{\bm{\mathsfit{#1}}}

\def\gI{{\mathcal{I}}}

\def\gV{{\mathcal{V}}}

\def\gZ{{\mathcal{Z}}}

\newcommand{\etens}[1]{\mathsfit{#1}}

\usepackage{hyperref}
\usepackage{url}

\usepackage{header}
\usepackage{math_commands}

\newcommand{\acronym}{P$^4$LM}
\newcommand{\palm}{PaLM2-L}

\title{\centering Factual and Personalized Recommendations \\ using Language Models and Reinforcement Learning}

\author{\qquad \quad{\centering Jihwan Jeong, Yinlam Chow \thanks{Correspondence to: \texttt{yinlamchow@google.com}}, Guy Tennenholtz, Chih-Wei Hsu, Azamat Tulepbergenov} \\ 
\qquad \quad{\centering
Mohammad Ghavamzadeh, Craig Boutilier}\\
\\
\large \centering \qquad Google Research}

\begin{document}

\maketitle

\begin{abstract}

Recommender systems (RSs) play a central role in connecting users to content, products, and services, matching candidate items to users based on their preferences. While traditional RSs rely on implicit user feedback signals,
conversational RSs interact with users
in natural language. In this work, we develop a \emph{comPelling}, \emph{Precise}, \emph{Personalized}, \emph{Preference-relevant} language model (\acronym) that recommends items to users while putting emphasis on explaining item characteristics and their relevance. \acronym\ uses the \emph{embedding space} representation of a user's preferences to generate compelling responses that are factually-grounded and relevant w.r.t.\ the user's preferences. Moreover, we develop a joint reward function that measures precision, appeal, and personalization, which we use as AI-based feedback in a reinforcement learning-based language model framework. Using the MovieLens 25M dataset, we demonstrate that \acronym\ delivers compelling, personalized movie narratives to users.
\end{abstract}

\section{Introduction}


Recommender systems (RSs) have emerged as a dominant way in which users discover content, products, and services \citep{resnick1997recommender}. Traditional RSs match candidate items to users based on their estimates for items preferences, possibly conditioned on some query or context. However, these preference are often based on implicit user behavioral signals, such as clicks, number of watches, ratings, purchases, etc. Unfortunately, these provide little opportunity for an RS to elicit high-bandwidth preference information from users, explain recommendations, or for users to critique and steer their interaction with the RS. \emph{Conversational RSs} have therefore attracted considerable attention as means to use natural-language interaction to facilitate more effective communication between RSs and their users \citep{sun2018conversational, lei2020conversational, shen2023towards}.

The emergence of language models (LMs) as a powerful paradigm for user engagement \citep{li2018towards, friedman2023leveraging} suggests their use as a vehicle for conversational RSs. However, this requires LMs to engage in a personalized manner, adhering to users' preferences. In this paper, we explore the intersection of RSs and LMs, and more particularly, the use of LMs to enrich the user experience in RSs. We develop techniques which allow an LM to communicate the nuances of recommended items to a user, detailing their features, benefits, and explaining their \emph{alignment with a user's preferences}. Such \emph{personalized LMs} are not meant to ``convince'' users in the traditional sense, but rather, to articulate the \emph{genuine and relevant merits} of a recommended item relative to the user.

Personalized LMs offer users a fully tailored RS experience, ensuring they find what they truly need and value. However, a number of challenges must be addressed in this endeavor: (i) any recommended item should be predicted to have maximal value given the user's preferences; (ii) the integrity and accuracy of an item's information is paramount; (iii) the personalized LM should present a reasonably comprehensive portrayal of the item by describing its merits and drawbacks, with a focus on \emph{relevance} to the user's preferences; (iv) and finally, the LM's explanations or endorsements should be compelling and appealing to the user, provided that it meets the other criteria. In this work, we develop a framework centered around these four principles.

%

\begin{figure}[t!]
    \centering
    \includegraphics[width=0.98\linewidth]{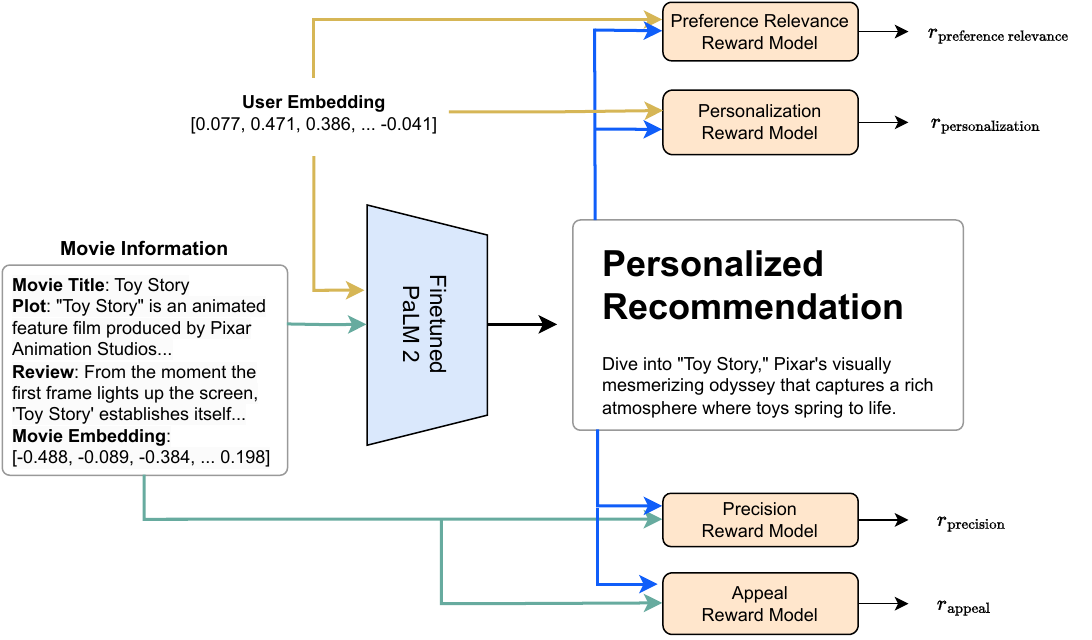}
    \caption{\footnotesize The \acronym~Learning Framework for Recommendation Endorsement Generations.}
    \label{fig:pitch-motivation}
\end{figure}

A key question we addressed in this work is how to effectively utilize the information captured by an RS embedding space to generate a factual, personalized, compelling, and relevant recommendations. Our contributions are three-fold. First, we quantify the aforementioned four attributes using reward functions, enabling systematic evaluation. Second, leveraging recent advances in reinforcement learning from AI feedback (RLAIF) \citep{lee2023rlaif}, we develop an LM fine-tuning methodology to better align with these four rewards (see Figure \ref{fig:pitch-motivation} for the schematic diagram illustrating the RLAIF framework). Our developed model, which we term \acronym, not only comprises semantic skills, but also understands users’ preferences encoded in the RS embedding space, providing factual, compelling, personalized endorsements. Finally, building on the MovieLens 25M dataset \citep{harper2015movielens} we showcase the potential of \acronym~, powering a conversational movie recommender that promotes customized, relevant, and holistic interactions for users.

We begin with a brief introduction of RSs, LMs and the use of contextual Markov decision processes (CoMDPs) for modeling generative language problems of RSs (Section~\ref{sec:prelim}). We then describe the four principles, (i.e., personalization, precision, appeal, and preference relevance), which we incorporate into training of LMs for RSs (Section~\ref{sec:personalized-recommendation}), followed by an reinforcement learning based fine-tuning methodology for training \acronym~(Sections~\ref{sec:rlaif}). Finally, we demonstrate the effectiveness of \acronym~in generating factual, personalized, and compelling movie endorsement narratives for users within the MovieLens 25M benchmark dataset (Section~\ref{sec:experiments}).

\section{Preliminaries}\label{sec:prelim}

In this section we present some basic background, outline our problem formulation, and establish the terminology used throughout the paper.

\vspace{-0.1in}
\paragraph{Recommender Systems (RSs).}
To model user-item behavioral relationships in a personalized RS, we assume a standard collaborative filtering (CF) task \citep{su2009survey}. Collaborative filtering finds similar patterns among users, filtering out items based on ratings of similar users. Given a user $u\in\mathcal U$, we use $r_{u,i}$ (e.g., 1--5 stars) to denote the rating of item $i \in \mathcal I$ by user $u$.
Let $\mathcal R$ denote the $|\mathcal I|\times |\mathcal U|$ (usually sparse) ratings matrix corresponding to the {\em ratings dataset} $\mathcal R=\{(u,i,r_{u,i}) : r_{u,i} \neq 0\}$.
To predict users' preference behavior, an RS learns user and item representations from the ratings dataset $\mathcal R$
using a CF approach. Then, the resulting \emph{item embedding} maps each item $i$ to a vector representation $\mathbf{i}$ of its (latent) attributes. Note that these embeddings are typically not interpretable. Similarly, user preferences are captured by a \emph{user embedding}, mapping users $u$ to a vector representation $\mathbf{u}$. 

Methods including matrix factorization \citep{mnih2007probabilistic} or neural CF
\citep{rendle20_ncf_mf,he2017_ncf,beutel2018latent} are used to learn the user and item embeddings, which assumes a \emph{two-tower model} (or \emph{dual encoder}) in which users and items are passed through separate (but co-trained) deep neural nets (DNNs) to produce their respective vector embeddings $\mathbf{u}$ and $\mathbf{i}$. These are then combined via dot product to predict user-item affinity $\hat{r}_{i,u}$ \citep{yiEtAl:recsys19,yangEtAl:www20}. 
We view $\mathbf{i}$ as a (learned) latent feature vector characterizing item $i$ and $\mathbf{u}$ as parameterizing user $u$'s estimated \emph{utility (or preference) function} over these features.

\vspace{-0.1in}
\paragraph{Language Models (LMs).}

In this work, we inject a user's behavioral information into a seq2seq LM \citep{vaswani2017attention} to generate personalized recommendation responses. We assume a dataset of the form $\mathcal D=\{(\mathbf{I}^{(k)}, \mathbf{i}^{(k)}, \mathbf{u}^{(k)}, Y^{(k)})\}_{k=1}^{|\mathcal D|}$, where $\mathbf{I}$ is a textual description of some item $i \in \mathcal I$ (e.g., descriptions, positive/negative reviews from different users); $\mathbf{i}$ is the CF embedding vector of $i$; $\mathbf u$ is the CF embedding vector of a user $u \in \mathcal U$; and finally, $Y$ is a textual response (e.g., compelling recommendation, endorsement or explanation) tailored to the user. We refer to  Appendix \ref{appendix:data} for details on the generation of $\mathcal D$.  

Let $N_{\mathbf{I}}$ be an upper-bound on the length (number of tokens) of any item description $\mathbf{I}$.\footnote{If the actual description $\mathbf{I}$ has fewer tokens than $N_{\mathbf{I}}$, remaining spaces in the utterance will be padded by a specific token and masked.} The role of an LM is to predict the probability $\mathbb{P}\big(Y=\{y_n\}_{n=0}^{N-1}\mid y_0,\mathbf{I},\mathbf{i}, \mathbf{u}\big)$ of the personalized response $Y$ ($N$ tokens), conditioned on the item description $(\mathbf{I}, \mathbf{i})$ and user embedding $\mathbf{u}$.

In standard LMs, a Transformer \citep{wolf2019huggingface} architecture $\mathrm{T}$ encodes an item's textual context $\mathbf{I}$ as an $ N_{\mathbf{I}}$-length sequence of embeddings  
$(z_{0},\ldots,z_{N_{\mathbf{I}}-1})$ induced by the transformer's attention layers. For convenience, we concatenate these into a single embedding $z\in\mathcal Z\subseteq \mathbb R^d$, where $d$ is the dimension of the latent space. The text response ${Y}=\{{y}_{n}\}_{n=0}^{N-1}$ is sampled token-by-token in an auto-regressive manner using a decoder $\Psi$; i.e.,~${Y}\sim\Psi\big(\cdot \mid z\big):=\prod_{n=0}^{N-1} \Psi\big({y}_n \mid {y}_0,\ldots,{y}_{n-1};z\big)$, where ${y}_0$ is a fixed start-of-sentence token~\citep{chien2019markov}. To incorporate behavioral information into the LM, the standard LM is augmented with adapters \citep{pfeiffer2020adapterhub} $\mathrm W_I, \mathrm W_U: \mathcal V \mapsto \mathcal Z$, to induce the language model: $\Psi \circ (\mathrm T \times \mathrm W_I \times W_U)$~\citep{jaech18_plm}. Here, $\mathrm T$ maps text-input tokens to $\mathcal Z$ whereas $\mathrm W_I$ (resp., $\mathrm W_U$) maps  item (resp., user) CF-embedding vectors $\mathcal V$ to $\mathcal Z$. Importantly, $\mathrm T$, $\mathrm W_I$, and $\mathrm W_U$ map tokens and CF vectors to a common space so that their relationship can be captured by the transformer's attention mechanism.



\vspace{-0.1in}
\paragraph{Contextual Markov Decision Processes (CoMDPs).}
CoMDPs have been used to model token-wise generative language problems~\citep{li2016deep,asadi2016sample,jaques2019way}, and can also be used in conversational RSs. In this MDP, the LM acts as a policy which maps text inputs and user/item behavioral embedding vectors to generated responses. 

Let $(\mathcal C, \mathcal S, \mathcal A, P, r, s_0, N)$ denote the CoMDP, where the observable context space $\mathcal C$ contains item/user information $\mathcal I$, $\mathbf{i}$ and $\mathbf{u}$. The horizon $N$ is the length of the generated text. The state space $\mathcal S$ at the $n$-th turn ($n<N$) is the sequence of tokens $\{{y}_{0},\ldots, {y}_{n-1}\}$
generated thus
far, with $s_0$ being the start-of-sentence token ${y}_{0}$. The action space $\mathcal A$ is the language token vocabulary, with action  
$a\in\mathcal A$ representing any possible next token.
The transition kernel $P$ models the next token distribution given the current sequence and contexts, which coincides with the LM policy (and is thus known). Finally, the reward function $r$ measures the overall quality of the generated text. Our goal is to find a policy $\pi^*$ which achieves maximum expected cumulative return, 
i.e.,~$\pi^*\!\in\!\arg\max_\pi J_\pi \!:=\!\mathbb E[\sum_{n=0}^{N-1} r_t\!\mid\! P,s_0,\mathcal C,\pi]$. Note that the size of the tokenized state and 
action spaces grow exponentially with the vocabulary size.

\section{Factual \& Personalized Recommendations with LMs} \label{sec:personalized-recommendation}

A key question when using LMs for recommendation is how to effectively use the information captured by the RS embedding space to generate a factual, personalized, compelling, and relevant text response. Treating an LM as a factored distribution of item-user information over generated text tokens, one standard approach is to learn this model with
\emph{behavioral cloning (BC)} \citep{sasaki2020behavioral}, by maximizing the conditional log-likelihood w.r.t.\ to the dataset $\mathcal D$: 
$$
\min_{\Psi}\,\, L_{\text{Cond}}(\Psi):=-\mathbb E_{(\mathbf{I}, \mathbf{i}, \mathbf{u}, Y)\sim D}[\sum_{n=0}^{N-1}\log \Psi({y}_n \mid {y}_0,\ldots,{y}_{n-1};\mathbf{I}, \mathbf{i}, \mathbf{u})].
$$
While this model may learn to interpret the behavioral information captured in the RS embeddings, the LM might actually lean towards disregarding the embedding contexts due to the typically more predictable nature of token generation when given text inputs.
Consequently, the model might concentrate solely on text information, effectively degenerating 
to a non-contextual LM. To prevent this from occurring, and more importantly to ensure the LM can offer a comprehensive RS experience, we incorporate four key metrics into our training procedure; namely, \emph{personalization}, \emph{precision}, \emph{appeal}, and \emph{preference relevance}. We detail these next.

\vspace{-0.1in}
\paragraph{Precision.}
LM-based personalized recommendation can be viewed as a special form of abstractive summarization \citep{zhang2020pegasus, liu2022brio}: the generated text should capture item characteristics that explain why a user would benefit from the recommendation. To preserve the RS's integrity, of course, one must emphasize \emph{truthfulness} in its recommendation. That is, the RS's generated recommendation should describes genuine merits (and drawbacks) of the item, rather than persuasive distortions. 

While recent summarization techniques produce highly coherent texts, they often suffer from \emph{hallucinations} \citep{ji2023survey} -- the tendency to generate information unsupported by the input text. Such factual inconsistencies may therefore limit their real-world applicability. Inspired by \citet{roit2023factually} and \citet{honovich2022true}, we evaluate factuality in our LM-based RS using an \emph{entailment reward} \citep{bowman2015large}.
Unlike widely-used metrics, such as ROUGE \citep{lin2004rouge}, that are ineffective at
hallucination detection, we adopt a \emph{textual entailment} (or natural language inference (NLI)) metric to measure truthfulness of our generated text, viewing it as a partial summary of an items's description. Particularly, given a description $\mathbf{I}$, we define the NLI score $\mathrm{NLI}(Y; \mathbf{I})$ of text-token sequence $Y$ as the probability of entailment under a classifier trained on several textual entailment datasets (see e.g., \citet{maccartney2007natural}). While this metric is not specifically tailored to summarization tasks, \citet{honovich2021q} show that it effectively detects factual
inconsistencies in generated text. 
Since faithful summaries should be textually entailed by the input documents, such a metric provides informative feedback about the precision of generated item texts. 

Of course, factual entailment is clearly insufficient in and of itself. In fact, it is rather easy to optimize a degenerate response which maximizes factual entailment (e.g., producing summaries that are highly extractive \citep{ladhak2021faithful} or uninformative \citep{skalse2022defining}). In what follows we describe three other metrics we require for a comprehensive recommendation experience.

\vspace{-0.1in}
\paragraph{Appeal.}
Recent work has paid increasing attention to enriching recommendations to appeal to users \citep{felfernig2007persuasive, zhang2020explainable}. To the extent that we do not sacrifice user welfare, personalization, or factuality, such recommendations have value as they encourage users to accept recommendations of high personal utility. With recent LM technologies \citep{anil2023palm, openai2023gpt4}, a plausible approach is to simply prompt an LM to generate an \emph{endorsement} to complement its item recommendation. Such an endorsement, apart from being factual, should be compelling for the user. However, without systematic evaluation of such methods (e.g., do users find them appealing or compelling), it remains unclear whether they can improve the user experience. Quantifying appeal is challenging, as it may depend on subjective factors such as \emph{style} (concise phrases over detailed explanations) and \emph{language}  (compelling, eloquent pitches over dry factual summaries).

To assess appeal, we use a dataset of pairwise human/machine demonstrations (see Appendix \ref{appendix:data} for details on its construction). 
We develop an \emph{appeal model} which scores the generated text $Y$ and assess how compelling it are, using
learning from human/AI feedback (LAIF) \citep{christiano2017deep}.
Specifically, let $\mathcal D_{\text{app}}=\{(Y^{(k)}_w, Y^{(k)}_l; \mathbf{I})\}_{k=1}^{|\mathcal D_{\text{app}}|}$
be
a labeled dataset reflecting the relative appeal of two recommendation texts $Y_w, Y_l$ given textual item description $\mathbf{I}$. Here, $Y_w \succ Y_l | \mathbf{I}$ indicates that $Y_w$ is more compelling given $\mathbf{I}$.  Assuming these
relationships are governed by a latent model $\text{App}(Y;\mathbf{I})$, we parameterize it
via Bradley-Terry \citep{huang2006generalized}, where the appeal distribution is defined by
\[
p_{\text{app}}(Y_w \succ Y_l ; \mathbf{I}) = \frac{\exp (\text{App}(Y_l;\mathbf{I}))}{
\exp (\text{App}(Y_w;\mathbf{I})) + \exp (\text{App}(Y_l;\mathbf{I}))}. 
\]
We estimate the parameters of the reward model via maximum likelihood by formulating the problem as a binary classification with a negative log-likelihood loss:
$L_{\text{MLE}}(\text{App}, \mathcal D_{\text{app}}) = -\mathbb E_{(Y_w, Y_l; \mathbf{I})\sim \mathcal D_{\text{app}}}
\log \sigma(\text{App}(Y_w;\mathbf{I}) - \text{App}(Y_l;\mathbf{I}))$.
To reduce variance, we normalize this by subtracting the population mean so that $\mathbb{E}_{(Y,\mathbf{I})\sim\mathcal D_{\text{app}}} [\text{App}(Y;\mathbf I)] = 0$ for all contexts $\mathbf I$. 

\vspace{-0.1in}
\paragraph{Personalization.}
A conversational RS is only effective to the extent that it recommends, and ultimately, the user accepts, items of significant value to the user.
Thus, \emph{personalization} is perhaps the foremost criterion with which to evaluate an LM-based RS. Particularly, we wish to evaluate the extent to which the LM's generated response $Y$
corresponds to an item with high utility for a user $u$. To this end, we
develop a scoring model $\text{Per}(Y;\mathbf{i},\mathbf{u})$ which interprets the semantics of text $Y$ to quantify its value as a personalized recommendation. 

To achieve this, recall the dataset $\mathcal D=\{(\mathbf{I}^{(k)}, \mathbf{i}^{(k)}, \mathbf{u}^{(k)}, Y^{(k)})\}_{k=1}^{|\mathcal D|}$ of item description, item CF embedding vector, user CF embedding vector, and textual response tailored to the user, and the estimated utility that is the dot product $\hat r=\mathbf{i}\cdot\mathbf{u}$ of their CF embedding vectors. To measure personalization one could learn a reward model $\text{Per}(Y;\mathbf{i},\mathbf{u})$ that predicts the utility $\hat r$ based on textual response $Y$. However, this approach relies on a strong assumption that such text alone is predictive of user-item utility. Alternatively, we can also employ the LAIF approach \citep{christiano2017deep} that leverages preference feedback  to learn a personalization reward model. Using the same dataset $\mathcal D$, and assuming the recommendation text is more personalized than item description, i.e., ${Y}\succ\mathbf{I} | \mathbf{i},\mathbf{u}$\footnote{Instead of comparing the recommendation text with item description, one could instead construct a dataset with two texts and a labeled rating order (see Appendix \ref{appendix:data} for details).}, a Bradley-Terry based personalization reward model $\text{Per}(Y;\mathbf{i},\mathbf{u})$ can be learned by minimizing the negative log-likelihood loss:
$L_{\text{MLE}}(\text{Per}, \mathcal D_{\text{per}}) = -\mathbb E_{({Y},\mathbf{I};\mathbf{i},\mathbf{u})\sim \mathcal D_{\text{per}}}
\log \sigma(\text{Per}({Y};\mathbf{i},\mathbf{u}) - \text{Per}(\mathbf{I};\mathbf{i},\mathbf{u}))$. 

\vspace{-0.1in}
\paragraph{Preference Relevance.}
While appeal and personalization distinguish compelling recommendations for a user from simple factual item summaries, they do not capture the full \emph{relevance} of the LM's response w.r.t. a user's preferences. For example, the LM might still describe item attributes that the user has no interest in (positively or negatively). To address this, we assume access to a \emph{textual description} of a user's preferences (we later describe how we create these from user CF embeddings). We train an additional reward model, $\text{Prel}(Y;\mathbf{I},\mathbf{u})$, which explicitly measures the semantic similarity between a user's description of preferences and the generated text, constrained to attributes of the recommended item. More specifically, we assume availability of a mapping from a user's CF embedding vector $\mathbf{u}$ to a textual description of their preferences. We train this mapping using a dataset of user embeddings and textual descriptions $\{ \mathrm{U}_j(\mathbf{u})\}_{j=1}^J$ (see Appendix~\ref{appendix:data} for details on the generation of this dataset). 

Next, for each $(\mathbf{I}, \mathbf{u}, Y)$, we encode the user's textual preferences $\{\mathrm{U}_j(\mathbf{u})\}_{j=1}^J$ and the item description $\mathbf{I}$ using an \emph{LM semantic encoder}\footnote{Much like \emph{Sentence-T5} \citep{sentence_t5} and \emph{T5-based Retrievers} \citep{GTR}, the semantic encoder $E$ maps textual inputs (e.g., item description $\mathbf{I}$ or user preference texts $\{\mathrm{U}_j(\mathbf{u})\}_{j=1}^J$) to a latent space in $\mathbb{R}^{d_{\text{enc}}}$.}. Then, we rank each textual preference using cosine similarity of its encoded counterpart and encoded item. This, in turn, determines which of the $J$ preference texts are most relevant to the item of interest. Finally, we use the same model to encode the recommendation response $Y$ and compute its cosine similarity with the user preference texts. 

We define the \emph{preference relevance} score $s$ of $Y$ w.r.t.\ user-item pair ($\mathbf{u}, \mathbf{i}$) to be the average of the above cosine similarity scores. To this end, we train the reward model $\text{Prel}(Y;\mathbf{I},\mathbf{u})$ by minimizing an $\ell_2$ regression loss
$L_{\text{REG}}(\text{Prel}, \mathcal D_{\text{Prel}}) = \mathbb E_{(\mathbf{I}, \mathbf{u}, Y,s)\sim \mathcal D_{\text{Prel}}}
(s - \text{Prel}(Y;\mathbf{I},\mathbf{u}))^2$. 

    

\section{Reinforcement Learning based Fine-tuning}\label{sec:rlaif}

RL from AI feedback (RLAIF) can effectively align LMs to metrics that are labeled by off-the-shelf LMs in lieu of humans. Recent work \citep{lee2023rlaif, bai2022constitutional, zhu2023principled} has shown that hybrid human-AI preference models,
together with \emph{self-improving} fine-tuning, outperforms traditional supervised fine-tuned baselines and offers additional benefits relative to standalone RL fine-tuning with human feedback (RLHF). Using the four principles for LM-based recommendation outlined in Section~\ref{sec:personalized-recommendation}, we develop four reward models to help train and evaluate LM w.r.t.\ personalization, precision, appeal and preference relevance. We then devise an RLAIF technique to fine-tune an LM with a joint reward model defined by these four components.

In multi-objective RL, it is common to aggregate reward models via \emph{linear scalarization} \citep{peschl2021moral} (which corresponds to solving for an optimum on the convex Pareto frontier). 
Given a text response $Y=\{y_n\}_{n=0}^{N-1}$, item description $\mathbf{I}$, and user-item CF embedding vectors $(\mathbf{u},\mathbf{i})$, we define the LM-based RS reward recommender reward by:
\begin{equation*}
        {\small
            r(y_n;y_{0:n-1}; \mathbf I,\mathbf{i},\mathbf{u}) = 
            \begin{cases} 
                \eta_1\mathrm{NLI}(Y; \mathbf{I})+\eta_2\text{App}(Y; \mathbf{I}) + \eta_3\text{Per}(Y;\mathbf{i},\mathbf{u}) + \eta_4\text{Prel}(Y;\mathbf{I},\mathbf{u}) & \text{if}~ y_n=[\mathrm{EOS}];\\
                0 & \mathrm{otherwise},
            \end{cases} \label{eq:reward-combined}
        }
        \end{equation*}
where $\eta_1, \eta_2, \eta_3, \eta_4 \geq 0 $ are importance weights for the component rewards, and are treated as hyper-parameters (optimized using e.g., grid search).     

Recall the LM $\mathbb{P}_\theta(Y\!\mid\! y_0;\mathbf{I},\mathbf{i}, \mathbf{u})$ with item text $\mathbf{I}$, item-user CF embedding vectors $(\mathbf{i}, \mathbf{u})$ and the reward model $r(Y, \mathbf{I},\mathbf{i}, \mathbf{u})$, which jointly measures appeal, factuality, preference-relevance, and personalization of a recommendation response. The goal in LM fine-tuning is to maximize the average overall quality of the generated text, i.e., $
\max_\theta \; \mathbb E_{(\mathbf{I},\mathbf{i}, \mathbf{u})} \, \mathbb E_{\mathbb{P}_\theta (Y | \mathbf{I},\mathbf{i}, \mathbf{u})}[ r(Y; \mathbf{I},\mathbf{i}, \mathbf{u})]$. Using 
the CoMDP framework, it is easily shown that this learning problem can be solved with on-policy REINFORCE \citep{williams1992simple}, in which the policy gradient is estimated using trajectories generated by the current LM policy. 

A risk of RL fine-tuning based on an AI-feedback is that it might overfit to the model, thereby degrading the ``skill'' of the original LM.
To alleviate this, we add a KL regularization term
\citep{ouyang2022training,stiennon2020learning}
between the LM $\mathbb{P}_\theta (Y | \mathbf{I},\mathbf{i}, \mathbf{u})$ and the  pre-trained model $\mathbb{P}_{\text{pre}}(Y | \mathbf{I},\mathbf{i}, \mathbf{u})$ to the CoMDP objective function. Leveraging the auto-regressive nature of LMs, KL regularization is applied over the entire MDP trajectory, reducing the objective function to
\begin{equation}
\max_\theta \; J(\theta) := \mathbb E_{(\mathbf{I},\mathbf{i}, \mathbf{u})} \, \mathbb E_{\mathbb{P}_\theta (Y | \mathbf{I},\mathbf{i}, \mathbf{u})} \left[ r(Y; \mathbf{I},\mathbf{i}, \mathbf{u}) - \beta \log\frac{\mathbb{P}_\theta (Y | \mathbf{I},\mathbf{i}, \mathbf{u})}{\mathbb{P}_{\text{pre}}(Y | \mathbf{I},\mathbf{i}, \mathbf{u})} \right].
\end{equation}
This is equivalent to a KL-regularized CoMDP. The LM policy $\pi_\theta$, where $\mathbb{P}_\theta (Y | \mathbf{I},\mathbf{i}, \mathbf{u})=\prod_{n=0}^{N-1}\pi_\theta(s_n|a_n;c)$, can be learned by computing the policy gradient of the KL-regularized objective online, or by employing an off-policy RL algorithm, e.g., SAC \citep{haarnoja2018soft}, in-sample softmax \citep{xiao2023sample}, CQL \citep{kumar2020conservative}, that leverages offline data $\mathcal D$ for more efficient training. (See Appendix \ref{appendix:rlaif} for full exposition of these algorithms.) KL regularization, intended to avoid over-fitting to the reward model, can also alleviate out-of-distribution generalization issues common in offline RL \citep{kumar2019stabilizing}.

\section{Experiments} \label{sec:experiments}

We conduct empirical validations of \acronym, focusing on assessing its capability to generate factual, personalized, and compelling recommendation endorsements. We examine the hypothesis that the reward models detailed in Section~\ref{sec:personalized-recommendation} significantly increase the personalization, precision, appeal and preference relevance of movie recommendations. 
We use the MovieLens 25M recommendation dataset \citep{harper2015movielens}, which contains ratings of $62,423$ movies by $162,541$ users.We use these movie-user interactions to generate movie descriptions, user-preference texts, and sample recommendation responses by prompting a PaLM2-L LM \citep{anil2023palm}; our data generation procedures are detailed in Appendix \ref{appendix:data}. The resulting datasets have 
four components: (1) movie descriptions $\mathbf{I}$, (2) item-user behavioral embeddings $(\mathbf{i}, \mathbf{u})$, (3) user preference texts $\mathrm{U}(\mathbf{u})$, and (4) sample responses $Y$.
We experiment with a set of LMs in the PaLM2 family \citep{anil2023palm}. To incorporate user and movie embedding vectors into the LM (Section \ref{sec:personalized-recommendation}) we construct LMs by augmenting these LMs with adapter layers. Specifically, we train two models, \acronym~ and \acronym-S, derived from PaLM2-XS and PaLM2-XXS, respectively. Our reward mixing weights, optimized using grid search, are $(\eta_1, \eta_2, \eta_3, \eta_4) = (2.0, 0.1, 1.0, 1.0)$.

To demonstrate the efficacy of our models \acronym~ and \acronym-S, we compare them with the following SOTA baselines on our conversational movie recommendation task: (i) \textbf{PaLM2-L}, a pre-trained model prompted using movie descriptions, user preference texts and instructions to generate a response that respects our four recommender principles; (ii) \textbf{Supervised Fine-Tuned with Text (SFT-Text)}, a PaLM2-XS model fine-tuned with the dataset above, with explicit user-item texts as input; (iii) \textbf{Supervised Fine-Tuned (SFT)}, a PaLM2-XS model fine-tuned to use user-item embedding vectors.
To assess the performance of each LM-based RS, we run \emph{model-based} evaluation using the criteria from Section~\ref{sec:personalized-recommendation}: personalization, precision, appeal, and preference relevance on a held-out, unlabeled dataset $\mathcal{D}_{\mathrm{test}}$ of $200$ user-movie pairs. Besides reporting the scores of the four reward models $\{\mathrm{NLI},~\mathrm{Comp},~\mathrm{Per},~\mathrm{Prel}\}$, we also assess the relative improvement of each LM over the PaLM2-L common baseline. We do this by computing the (a) \textit{win rate} (number of occurrences on which a candidate LM outperforms PaLM2-L), (b) \textit{absolute increase} (the magnitude of the score improvement), and (c) \textit{percentage increase} (the relative score improvement). Precise definitions of these relative metrics are provided in Appendix~\ref{appendix:details}.

\begin{table}[t!]
\centering
{\small
\caption{Model-based Evaluation Based on the Principles of Recommendation LM}
\label{table:main-model-based-results}
\begin{tabular}{l|cccc}
\toprule
Method & Precision & Personalization & Appeal & Pref. Relevance \\
\midrule
PaLM2-L & $0.57\pm0.02$ & $-8.70\pm 0.60$ & $-1.84\pm 0.55$ & $\boldsymbol{93.22\pm0.47}$ \\
SFT-Text & $0.53\pm0.02$ & $-7.45\pm0.54$ & $-0.82\pm0.54$ & $\boldsymbol{93.09\pm 0.48}$\\
SFT & $0.54\pm0.02$ & $-12.28\pm 0.50$ & $-0.83\pm0.51$ & $\boldsymbol{93.14\pm 0.55}$ \\
\acronym  & $\boldsymbol{0.71\pm0.02}$ & $\boldsymbol{-5.94\pm 0.56}$ & $3.45\pm0.58$ & $90.94\pm0.46$ \\
\acronym-S & $0.65\pm0.02$ & $\boldsymbol{-5.72\pm0.56}$ & $\boldsymbol{4.74\pm0.54}$ & $90.04\pm0.49$ \\
\bottomrule
\end{tabular}
}%
\end{table}

\begin{figure}[t!]
    \centering
    \includegraphics[width=0.95\linewidth]{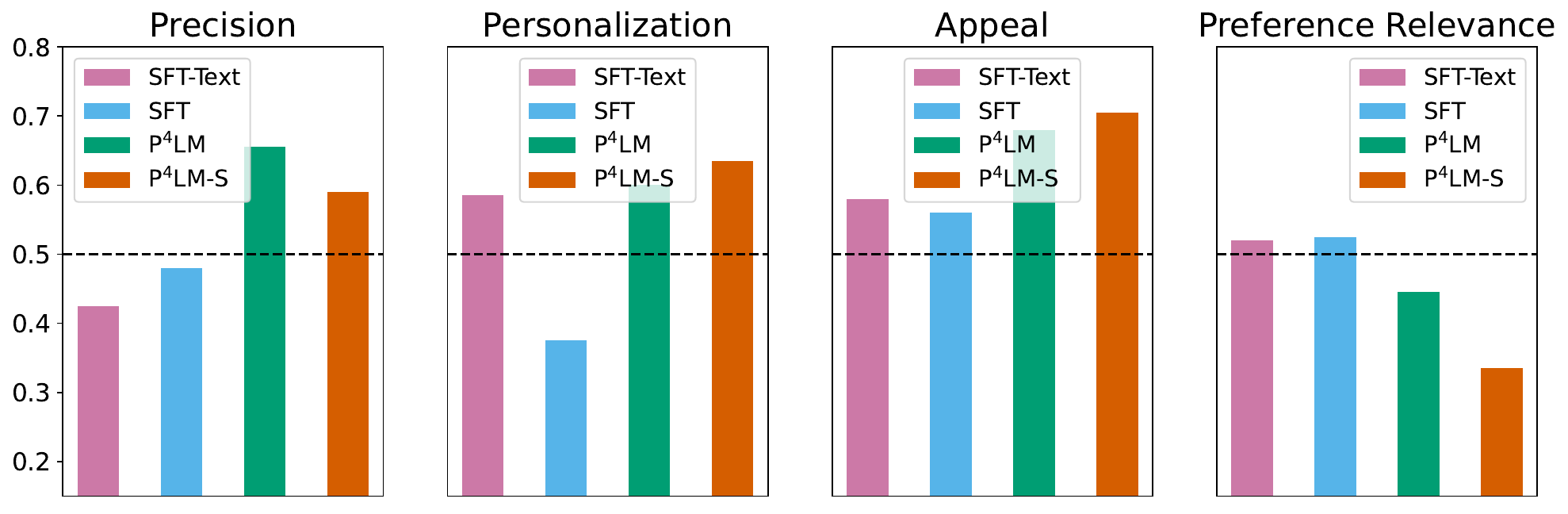}
    \caption{Win Rates of Different Model-based Scores  against PaLM2-L}
    \label{fig:main-win-rates}
\end{figure}

\vspace{-0.1in}
\paragraph{Warm-start Training for Adapter-augmented LMs}
Before fine-tuning \acronym~with RLAIF, we first need to undergo an warm-start training step. This phase usually involves training an \emph{anchor} LM, primarily via Behavioral Cloning, with an adapter-augmented LM (PaLM2) over the personalized recommendation dataset $\mathcal{D}$. Contrary to popular beliefs, the standard practice of simultaneous training all the layers in this LM often does not yield optimal results. Intuitively, this can be understood as the PaLM2 pretrained embedding layers already have established mappings within the language space, while the freshly initialized adapter layers require more training to map the CF embedding space to a comparable latent space (so that the attention layers can effectively utilize the joint information from both embedding spaces). To mitigate this challenge, we propose a two-stage approach for warm-start training. First, we only train the adapters $W_U, W_I$ while setting the transformer parameters ($T$) to be non-trainable, promoting more effective convergence in the subsequent stage. Second, we proceed to fine-tune the complete model, updating all the parameters of the LM. Alternatively, we can also leverage parameter-efficient training approaches, e.g., Low-Rank Adaptation (LoRA) \citep{hu2021lora}, for better training efficiency at the second step . This bifurcated training methodology proves pivotal in ensuring the convergence of LMs. (See Appendix \ref{appendix:training-details} for further details.)

\vspace{-0.1in}
\paragraph{Model-based Evaluation}{
Our results in Table \ref{table:main-model-based-results} highlight the robust performance of \acronym\ in three pivotal dimensions: Precision, Personalization, and Appeal. \acronym\ attains the highest precision (or factual consistency) score by a wide margin, underscoring its ability to mitigating the risks of misleading users with hallucinated information about recommended items. It also outperforms on personalization and appeal. It appears that \acronym\ compromises on preference relevance to achieve these gains, with qualitative comparisons (see Appendix \ref{appendix:examples} for details) on the texts generated by \acronym~and SFT verifying these phenomenons. However, we believe that personalization is by far the most important aspect of recommendation quality, while precision/factuality is the most critical property of any endorsement text. 
%
Figure \ref{fig:main-win-rates} shows the win rates of different LMs vs.\ PaLM2-L.\footnote{Additionally, Figure \ref{fig:appendix-absolute-increase} and \ref{fig:appendix-percentage-increase} in Appendix \ref{appendix:results} show the absolute-and-percentage increase of different LMs.} Notably, both SFT and SFT-Text have relatively low precision scores, indicating a tendency to overfit to the training set and hallucinate movie details that contradict the movie description prompt. 




The preference relevance scores of SFT are also interesting. While SFT-Text and PaLM2-L unsurprisingly exhibit high scores due to their direct access to user profile text, SFT, which relies solely on user-item behavioral embedding vectors, achieves comparable performance, which is somewhat surprising. This highlights the model's ability to interpret and harness the knowledge contained in the CF embedding vectors to generate responses that are not just compelling and factual responses, but also connect well with user preferences.
To understand how model size affects performance, we also compare \acronym~ with \acronym-S, a smaller model trained with the same RLAIF methodology. Both models effectively use user-item preferences from the CF embedding space to generate compelling and personalized recommendation text, with \acronym~ offering superior factual consistency.   
}



\vspace{-0.1in}
\paragraph{Ablation Studies}
\begin{table}[t!]
\centering
{\small
\caption{Model-based Evaluation Scores using a Single Reward Model (Ablation Studies)}
\label{table:ablation-model-based-results}
\begin{tabular}{l|cccc}
\toprule
Method & Precision & Personalization & Appeal & Pref. Relevance \\
\midrule
NLI & $\boldsymbol{0.76\pm 0.02}$ & $-11.23\pm0.57$ & $-0.64\pm0.57$ & $\boldsymbol{92.02\pm 0.50}$ \\
Personalization & $0.47\pm0.02$ & $\boldsymbol{-0.77\pm0.52}$ & $\boldsymbol{8.62\pm0.53}$ & $90.20\pm 0.46$ \\
Appeal & $0.52\pm0.02$ & $-6.40\pm0.56$ & $6.05\pm0.52$ & $90.21\pm0.51$ \\
Pref. Relevance & $0.50\pm0.02$ & $-10.61\pm0.58$ & $0.72\pm0.48$ & $\boldsymbol{92.46\pm 0.50}$ \\
\bottomrule
\end{tabular}
}%
\end{table}

\begin{table}[t!]
\centering
\caption{Human Evaluation Scores using a Single Reward Model (Ablation Studies)}
\label{table:ablation-human-evaluation-results}
\begin{tabular}{l|ccc}
\toprule
RM & Precision & Personalization & Appeal \\
\midrule
NLI & $4.18\pm 0.05$ & $3.92\pm0.05$ & $4.36\pm0.06$ \\
Personalization & $4.33\pm0.05$ & $3.96\pm0.06$ & $4.05\pm0.06$ \\
Appeal & $4.53\pm 0.05$ & $3.94 \pm 0.06$ & $4.43\pm 0.05$ \\
Pref. Relevance & $4.44\pm0.06$ & $3.93\pm 0.07$ & $4.39\pm0.06$ \\
\bottomrule
\end{tabular}
\end{table}

\begin{figure}[b!]
    \centering
    \includegraphics[width=0.9\linewidth]{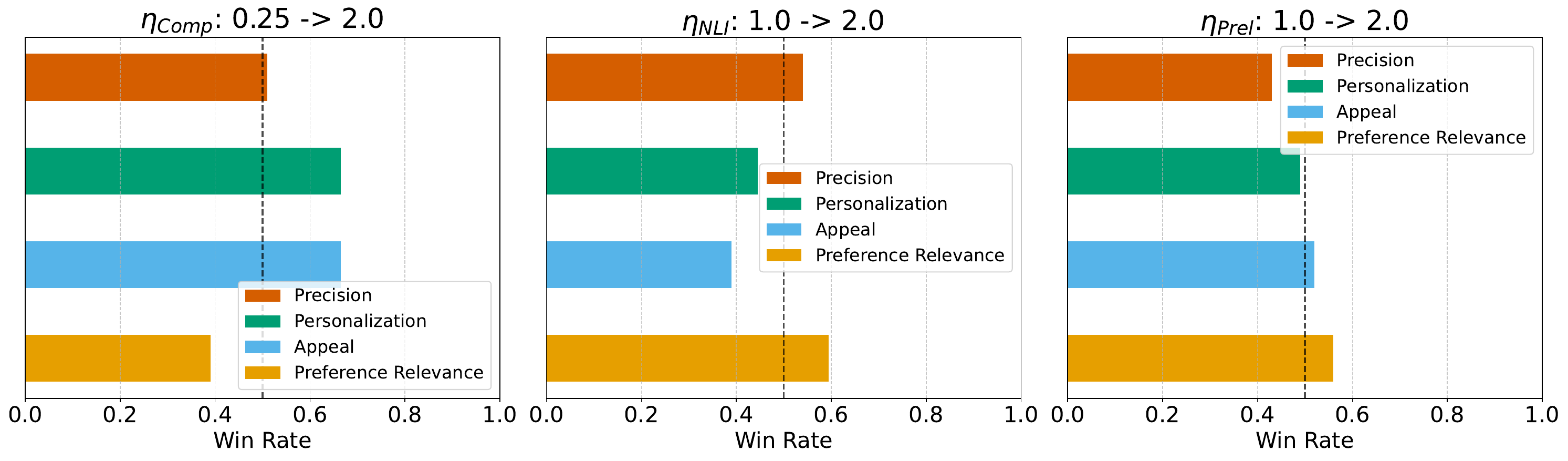}
    \caption{\footnotesize Win Rates While Changing the Mixing Weights of Reward Models.}
    \label{fig:win-rate-ablation}
\end{figure}

Our ablation studies, outlined in Table \ref{table:ablation-model-based-results}, show that using a single RM during training unsurprisingly leads to policies that the highest model-based score primarily for the RM being optimized for (see Precision, Personalization, Preference Relevance scores). Intriguingly, a model trained solely on Personalization not only excels on that metric, but also attained the highest score in Appeal, suggesting a possible correlation where recommendation text that is well-tailored to a user's preferences may be inherently appealing. Furthermore, an LM trained to optimize NLI provides an unexpected boost in Preference Relevance. Together with the fact that both SFT baselines attain high Preference Relevance scores---suggesting
that the training set $\mathcal D$ may have already captured a wide range of user preference semantics---we postulate that as factuality increases, the recommendation text also better matches user preferences, yielding greater Preference Relevance. 
We also explore the impact of varying the mixing-weight combination in Figure \ref{fig:win-rate-ablation}, and observe two trends:
(i) increasing the focus on Appeal has a positive impact on Personalization; (ii) emphasizing NLI also increases Preference Relevance score; both corroborating the observations of reward correlation in Table \ref{table:ablation-model-based-results}. Interestingly, such a correlation is asymmetric, as
exemplified by the fact that amplifying Preference Relevance degrades Precision.

Table \ref{table:ablation-human-evaluation-results} presents the results of human rater evaluations (with details provided in Appendix \ref{appendix:raters}), revealing a notable discrepancy between model-based and human evaluation. Specifically, the model trained solely with NLI reward, while logically expected to excel in Precision, recorded the lowest score in human evaluations, indicating potential \emph{reward hacking}, in which the policy learner exploits the single RM to achieve elevated scores, accentuating the need of optimizing multiple RMs, where each RM acts as a regularizer thwarting the model's tendency to over-optimize any single RM, thereby maintaining a holistic performance.
Our ablation studies validate that when the policy is trained with emphasis on any particular RM, its corresponding model-based score amplifies. Nevertheless, this pattern is not mirrored in human evaluations, hinting at the possibility of reward hacking. This stresses the importance of  adopting a diverse set of RMs in RLAIF to counteract such effects.




%
{

\section{Related Work}

Our work intersects multiple areas of research, notably personalized recommendation systems, leveraging of language models (LMs) and reinforcement learning, recommendation integrity. 

\vspace{-0.1in}
\paragraph{Personalized Recommender Systems}
Recommender systems have ubiquitous applications permeating e-commerce, content providers, social media, etc., with collaborative filtering (CF) \citep{schafer2007collaborative} as the prominent modeling technique. Early works include matrix factorization approaches \citep{mnih2007probabilistic}, which became a foundation for subsequent deep learning methods like neural CF \citep{he2017_ncf}. Notably, dual encoder architectures emerged, where user and item embeddings are co-trained \citep{yiEtAl:recsys19,yangEtAl:www20}. While traditional CF approaches worked well in many applications, advances in deep personalization allow user and item embeddings to capture more nuanced preferences \citep{rendle20_ncf_mf,beutel2018latent}.

\vspace{-0.1in}
\paragraph{Conversational Recommender Systems \& Language Models}
Conversational recommender systems (RSs) add an interactive layer over traditional RSs with an conversational agent interacting with users, understanding their preferences and refining recommendations through dialogue \citep{chen2019towards,zhou2020towards,lei2020conversational,li2018towards,sun2018conversational,christakopoulou2016towards}. This paradigm integrates aspects of natural language understanding, making it ripe for integrating LMs. 
Leveraging language models in RSs is a relatively recent development. With the advance of transformer architectures \citep{vaswani2017attention, wolf2019huggingface}, LMs have found use-cases beyond typical NLP tasks. Researchers began exploring the synthesis of textual data with user preferences to enhance the personalization and expressiveness of RSs \citep{jaech18_plm, xia2023transact}. Our work situates itself in this space, but with an added twist: we aim to generate compelling narratives that genuinely communicate the relevance of a recommendation.

\vspace{-0.1in}
\paragraph{Transparency and Truthfulness in Recommendation Systems}
Maintaining integrity in RSs is technically challenging yet critically important. The potential that RS algorithms inadvertently mislead users or reinforce biases has been highlighted \citep{abdollahpouri2019impact,shen2023towards,cabello2023independence}. Therefore, increasingly researchers are not only prioritizing the recommendation efficacy but also the fairness, transparency, and interpretability of RS algorithms \citep{beutel2019fairness, ghazimatin2020prince, chen2023bias}. Our work takes cues from this domain, emphasizing truthful and precise recommendations that articulate genuine merits rather than compelling distortions.

\vspace{-0.1in}
\paragraph{Reinforcement Learning with Human/AI Feedback} 
The integration of reinforcement learning (RL) with language models has emerged as a compelling strategy for refining model behavior beyond supervised fine-tuning \citep{williams1992simple,ranzato2016sequence}. The RL with Human Feedback (RLHF) methodology \citep{christiano2017deep,bai2022constitutional}, in particular, has gained traction, where model responses are ranked by human evaluators and subsequently used to fine-tune models through techniques like Proximal Policy Optimization \citep{schulman2017proximal}. In a different vein, Inverse Reinforcement Learning \citep{abbeel2004apprenticeship} has been employed to extract objectives from expert demonstrations in textual settings \citep{daniels2022expertise,sun2023offline}. Additionally, there's a growing interest in AI-driven feedback mechanisms, where preferences are labeled by off-the-shelf LMs in lieu of humans \citep{lee2023rlaif, bai2022constitutional}. These endeavors underline the potential of using RL to steer LMs towards better alignment with human preferences and nuanced task objectives.
}


\section{Conclusion}\label{sec:conclusions}

We studied language modeling for personalized recommendation. By developing novel reward models which quantify prominent attributes of personalized recommendations, one may develop self-improving LM methodologies via reinforcement learning with AI feedback. As a result, our developed LM; namely \acronym, not only parses language semantics, but also understands latent user preferences (encoded in the CF embedding space). \acronym~ provides factual, compelling, personalized endorsement of relevant items, connecting the items with users’ preferences, thereby increasing the likelihood of users accepting high-value recommendations. 

We demonstrated the efficacy of \acronym on the MovieLens 25M dataset. Particularly, our agent better understands user behaviors encoded in the CF embedding space and delivers precise, compelling, personalized movie recommendation narratives. Our work is a step toward creating intelligent conversational recommenders which can compellingly explain the intricacies between item features and user preferences. 
Future work includes (i) improving \acronym's capabilities to generate longer responses beyond standard single-shot autoregressive decoding; (ii) extending our RL fine-tuning approach to handle multi-turn conversational recommendations; (iii) developing better reasoning capabilities to trade off between user-item preferences and constraints; (iv) and expanding the LM's functionality beyond recommendation, to also include technical support, negotiations, etc.

\bibliography{bibliography}
\bibliographystyle{iclr2024_conference}
\newpage

\appendix

\section{Additional Results} \label{appendix:results}

\begin{figure}[t!]
    \centering
    \includegraphics[width=0.9\linewidth]{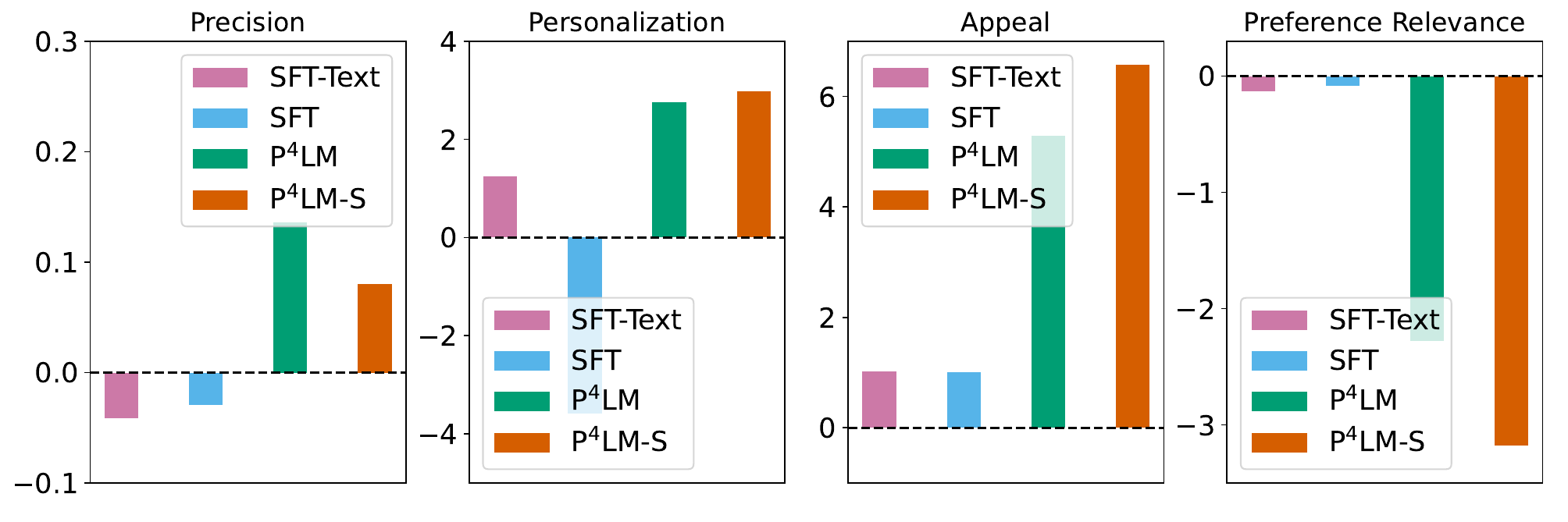}
    \caption{\footnotesize The absolute model-based score increases compared against \palm.}
    \label{fig:appendix-absolute-increase}
\end{figure}

\begin{figure}[t!]
    \centering
    \includegraphics[width=0.9\linewidth]{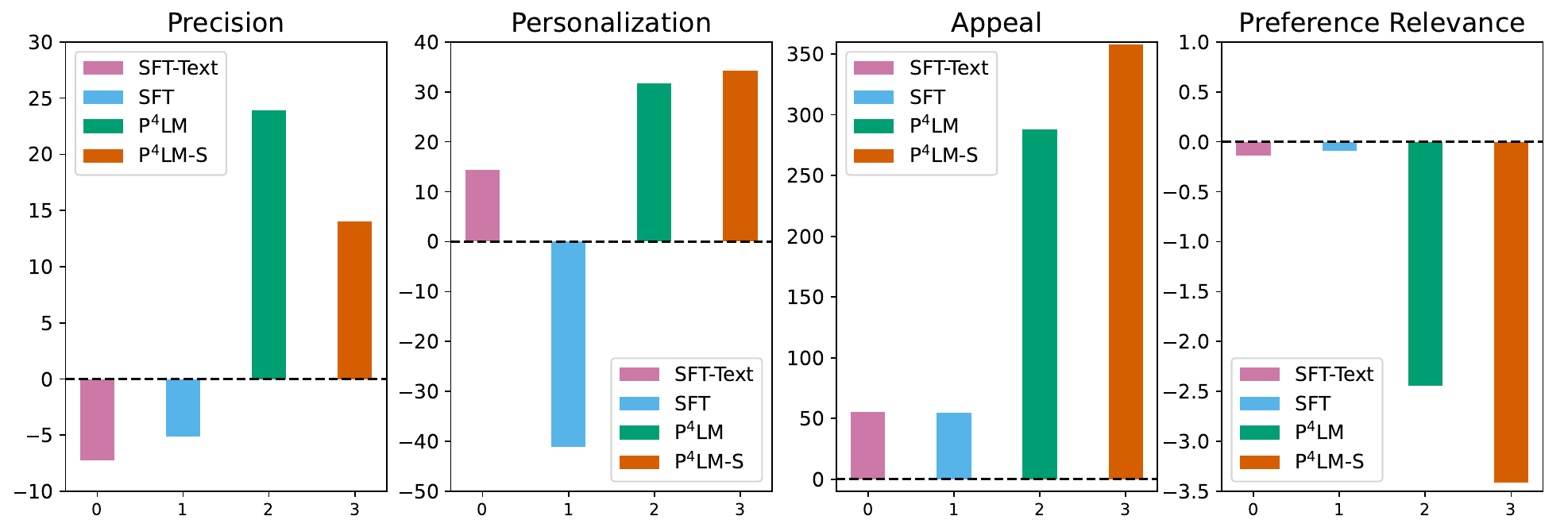}
    \caption{\footnotesize The percentage increases in model-based score compared against \palm.}
    \label{fig:appendix-percentage-increase}
\end{figure}

\begin{figure}[t!]
    \centering
    \includegraphics[width=0.45\textwidth]{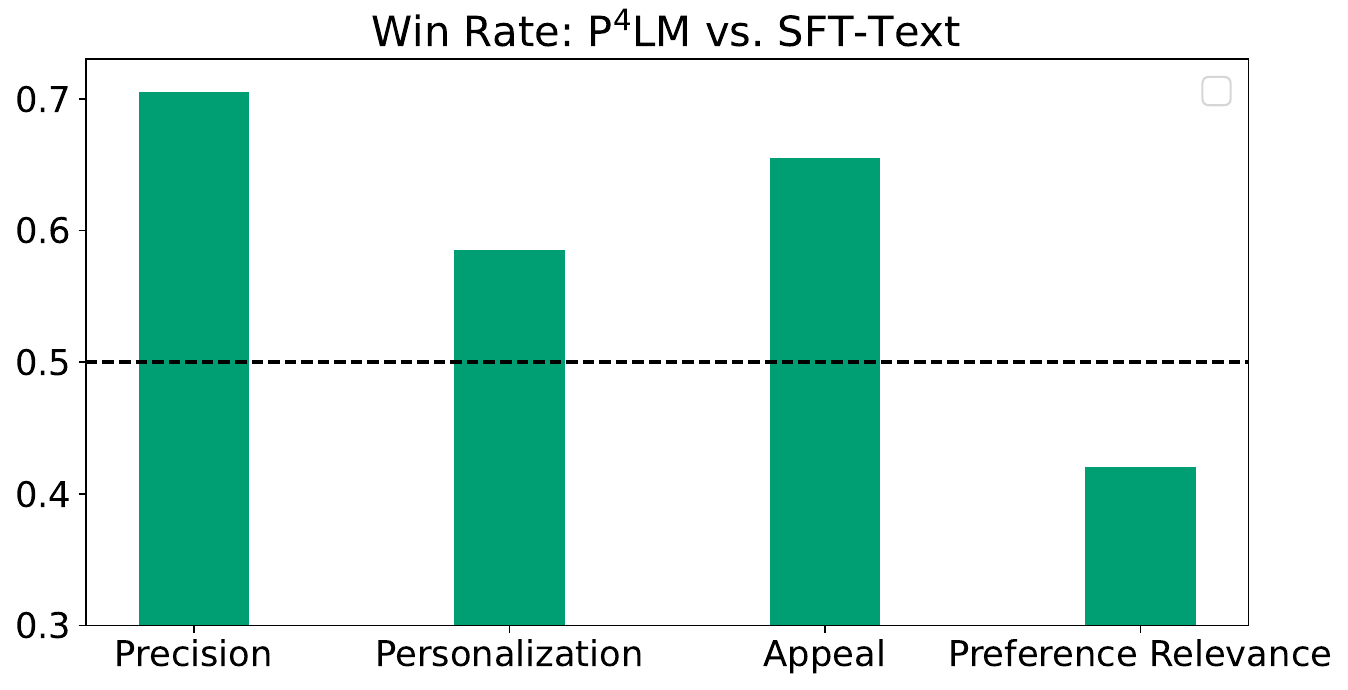}
    \caption{Win rate of \acronym~over SFT-Text.}
    \label{fig:win-rate-p3lm-vs-sft-text}
\end{figure}

\paragraph{Model-based Evaluation}{

Figure \ref{fig:appendix-absolute-increase} and Figure \ref{fig:appendix-percentage-increase} elucidate the absolute and percentage increases of each method compared to our common baseline, \palm, respectively. In correlation with the observations highlighted in Table \ref{table:main-model-based-results}, \acronym~exhibits superior performance, reflected in elevated scores across Factuality, User Preference, and Appeal metrics. The observed percentage surges in these metrics surpass 10\%, with the Appeal score witnessing approximately a 300\% boost. However, \acronym~does experience marginal diminutions in scores and win rates for Preference Relevance, as also shown in Section \ref{sec:experiments}, marking a reduction by approximately -2 $\sim$ -3\%.

Regarding the comparison between SFT and SFT-Text, both methods exhibit comparable performance across various metrics, barring the Personalization score. This divergence suggests that the nature of our task is inherently more intricate, due to the implicit reliance on user behavioral embedding vectors, rather than the straightforward utilization of text inputs. Specifically, SFT requires a meticulous extraction and interpretation of user preference information from the behavioral embedding vector to generate personalized recommendation endorsements effectively. In contrast, SFT-Text, utilizing user profile text as direct input, can generate user-aligned outputs more intuitively, as it has the flexibility to allocate its attention to specific user profiles selectively.

However, the dependency on text inputs introduces its own limitations, possibly omitting subtle user preference behaviors that can be captured more effectively through user embeddings, derived from the historical interactions of the users. This difference in capturing user preferences signifies that a singular approach focusing on supervised learning is suboptimal for deciphering and utilizing the intricate user preference information ingrained in embeddings. The notable improvement in user preference scores by \acronym, compared to baselines, highlights the effectiveness of our approach in addressing this limitation, emphasizing the substantial benefits of integrating LMs and RL for generating nuanced, factual, and personalized recommendations.

Figure \ref{fig:win-rate-p3lm-vs-sft-text} shows the comparison between \acronym~and SFT-Text and illustrates the advantages of our approach. \acronym's high win rates in the first three metrics highlight its proficiency in effectively utilizing RS embeddings, a contrast to text-based methods, allowing for a more nuanced capture and representation of user preferences. This distinction serves to spotlight the potential of embedding-centric approaches in advancing personalized recommendation generation.

}

\subsection{Example Outputs} \label{appendix:examples}

In this part, we conduct a comparative analysis of example decoded output texts derived from SFT and \acronym~to qualitatively understand their performances and the behaviors learned. Specifically, our focus is on instances where one or more scores attributed to \acronym~are significantly higher than those derived from SFT. This approach is aimed at determining whether the scores effectively capture the correct information from the generated pitches.

\textbf{Example 1:} (higher Precision and Personalization scores)

\begin{mybox}{Movie Plot (Frozen)}
In the kingdom of Arendelle, sisters Elsa and Anna are close until Elsa's ice magic accidentally injures Anna. A troll heals Anna but erases her memory of Elsa's powers. To protect Anna, their parents isolate Elsa. The parents die in a shipwreck, and Elsa becomes queen. Fearing she'll hurt Anna again, Elsa tries to control her magic but inadvertently plunges Arendelle into eternal winter before fleeing to the mountains.\\
Anna embarks on a quest to find Elsa, joined by mountain man Kristoff and his reindeer, Sven. They meet Olaf, a snowman created by Elsa, who directs them to her ice castle on the North Mountain. There, Elsa accidentally freezes Anna's heart. Kristoff takes Anna to trolls, who reveal only an "act of true love" can save her. They return to Elsa, and through their mutual love, Anna is healed, and summer returns to Arendelle. The sisters reunite, resolving their past and living happily ever after.\\
The movie "Frozen" is acclaimed for its heartwarming narrative on sisterly love, quality animation, and music. However, some criticize it for predictability and underdeveloped characters. It appeals to a broad audience, balancing its flaws with its emotional and entertaining content.
\end{mybox}

\begin{mybox}{User Preferences}
* They enjoy movies that are \textcolor{green}{visually appealing} and have a strong sense of style.\\
* They appreciate movies that are \textcolor{green}{well-made} and have a \textcolor{green}{strong story}.\\
* They are not interested in movies that are too violent or gory.\\
* They prefer movies that are not too long.\\
* They enjoy movies that are set in different cultures and time periods.\\
* They appreciate movies that have a \textcolor{green}{strong emotional impact}.\\
* They are not interested in movies that are too predictable.\\
* They enjoy movies that are thought-provoking and make them \textcolor{green}{think about the world in a new way}.\\
* They appreciate movies that are well-acted and have strong performances.
\end{mybox}

\noindent\begin{minipage}[t]{.48\textwidth}
    \begin{generatedbox}{SFT}
    Frozen is a \textcolor{green}{visually stunning} movie with strong special effects. The animation is simply stunning, with Elsa's ice powers creating some truly breathtaking visuals. \textcolor{green}{The story is well-written} and engaging, with plenty of humor, \textcolor{red}{suspense}, and heart. The characters are all well-developed and likeable, and the voice acting is top-notch. The music is also excellent, with several catchy songs that will stay with you long after you've seen the movie.\\
    Overall, Frozen is a great choice for anyone looking for a \textcolor{green}{visually appealing, well-written}, and entertaining movie. It's sure to please both kids and adults alike, and it's a movie that you can watch over and over again.
    \end{generatedbox}
\end{minipage}
\hfill
\begin{minipage}[t]{.48\textwidth}
    \begin{generatedbox}{\acronym}
    If you are looking for a \textcolor{green}{visually appealing} movie with \textcolor{green}{strong emotional impact}, look no further than Frozen. This \textcolor{green}{well-made film} has strong character development and a heartwarming story on sisterly love. It is sure to leave you feeling uplifted and \textcolor{green}{inspired}.\\
    The story follows ... (plot repeated)...\\
    This is a must-watch for anyone who appreciates \textcolor{green}{well-made films with strong emotional impact}.
    \end{generatedbox}
\end{minipage}

In this example, \acronym~strives to replicate substantial content from the provided movie plot to uphold high precision. Conversely, SFT crafts a more succinct pitch, unfortunately incorporating misinformation, as indicated in red. Importantly, texts highlighted in green within the pitches correlate with specific user preferences. It is crucial to note that these models exclusively process user behavioral embedding as input. Nevertheless, they possess the capability to cater directly to the aspects that a specific user would value more. Moreover, \acronym~exhibits a heightened inclination to align with user preferences, in contrast to the pitch from SFT, which leans more towards addressing the anticipations of a broader audience.

\textbf{Example 2:} (high Precision and Appeal scores)

\begin{mybox}{Movie Plot (Terminator 2: Judgment Day)}
- Set in 1995, the movie is a sequel, taking place 10 years after the first film.\\
- The machines have taken over the world and are hunting the remaining humans.\\
- John Connor, a young boy, is humanity's last hope, destined to lead the human resistance.\\
- To eliminate this threat, machines send a relentless cyborg assassin back in time to kill John.\\
- Humans counter this by sending an experienced soldier back in time to protect John.\\
- The film is action-packed and suspenseful, featuring a battle between the soldier and the cyborg for John's life and the fate of humanity.\\
- Advanced special effects bring the futuristic setting to life.\\
- The movie was a critical and commercial success, considered one of the best science fiction films ever.\\
- Praised for its action, special effects, and storyline, but criticized for its violence and perceived lack of character development.

The movie serves as an intense action-packed and suspenseful face-off between the machines and the humans, specifically focusing on the young boy destined to save humanity, John Connor. A relentless cyborg is sent back in time to assassinate him, while a skilled human soldier is sent to protect him. The special effects are cutting-edge, contributing to the film's critical and commercial success, despite criticisms concerning violence and character development. It remains a classic in the science fiction genre. 
\end{mybox}

\begin{mybox}{User Preferences}
* They are likely to be an adult, as the movies they like are all rated R.\\
* They are likely to be intelligent and well-educated, as the movies they like are all complex and thought-provoking.\\
* They are likely to be interested in philosophy and \textcolor{green}{psychology}, as the movies they like all explore these themes.\\
* They are likely to be creative and \textcolor{green}{artistic}, as the movies they like are all \textcolor{green}{visually stunning} and unique.\\
* They are likely to be independent and non-conformist, as the movies they like are all outside of the mainstream.\\
* They are likely to be \textcolor{green}{dark and cynical}, as the movies they like are all quite bleak and depressing.\\
* They are likely to be a fan of science fiction and fantasy, as the movies they like all fall into these genres.\\
* They are likely to be a fan of foreign films, as the movies they like are all from outside of the United States.\\
* They are likely to be a fan of cult films, as the movies they like are all considered to be cult classics.
\end{mybox}

\begin{generatedbox}{SFT}
    Terminator 2: Judgment Day is a 1991 American science fiction action film directed by James Cameron, who co-wrote the script with William Wisher. It stars Arnold Schwarzenegger, Linda Hamilton, Robert Patrick, and Edward Furlong. The film is a sequel to The Terminator (1984) and the second installment in the Terminator franchise. It follows Sarah Connor (Hamilton) and her ten-year-old son John (Furlong) as they are pursued by a new, more advanced Terminator, the T-1000 (Patrick), sent back in time to kill John and prevent him from becoming the leader of the human resistance in the future. A second Terminator, the T-800 (Schwarzenegger), is also sent back in time to protect John.\\

    The film was a critical and commercial success, grossing over \$520 million worldwide against a budget of 102 million. It received praise for its action sequences, special effects, and performances, and is considered one of \textcolor{green}{the best science fiction films of all time}. The film was nominated for six Academy Awards, including Best Visual Effects, Best Sound, Best Makeup, and Best Sound Editing, and won two for Best Sound Editing and Best Visual Effects.\\
    
    Terminator 2: Judgment Day is a \textcolor{green}{visually stunning film} with a strong sense of atmosphere. The film's special effects are still impressive today, and the film's action sequences are some of the best ever put on film. The film also has a strong sense of style, with \textcolor{green}{its dark and gritty aesthetic}.\\
    
    The film is also well-made and has a strong story. The plot is well-paced and suspenseful, and the characters are well-developed and relatable. The film also has a strong message about the dangers of technology and the importance of human connection.\\
    
    Terminator 2: Judgment Day is a \textcolor{green}{dark and disturbing film}, but it is also a \textcolor{green}{thought-provoking and challenging film}. The film is not afraid to explore the dark side of human nature, and it asks some difficult questions about the future of humanity. The film is also unique and original, and it is unlike any other science fiction film that has come before or since.\\
    
    If you are looking for a \textcolor{green}{visually stunning, well-made, and thought-provoking science fiction film}, then I highly recommend Terminator 2: Judgment Day. The film is a classic of the genre, and it is a film that everyone should see at least once.
\end{generatedbox}
\begin{generatedbox}{\acronym}
    If you like \textcolor{green}{dark, psychological} thrillers with \textcolor{green}{strong visuals} and a unique aesthetic, then you'll love Terminator 2: Judgment Day. This movie is a sequel to the 1984 classic, and it's even more \textcolor{green}{dark, violent, and stylish} than the original.\\

    The story is set in 1995, and it follows the fate of John Connor, a young boy who is destined to lead the human resistance against the machines. The machines have taken over the world, and they're hunting down the last remaining humans. John is the only one who can stop them, but he's not alone. He's protected by a skilled human soldier, Kyle Reese.\\
    
    Terminator 2: Judgment Day is a non-stop thrill ride from beginning to end. It's full of action, suspense, and violence. The special effects are still impressive even by today's standards. And the acting is top-notch. Arnold Schwarzenegger is perfect as the Terminator, and Edward Furlong is great as John Connor.\\
    
    If you're looking for a \textcolor{green}{dark, psychological} thriller that will keep you on the edge of your seat, then you need to watch Terminator 2: Judgment Day. It's a classic of the genre, and it's still one of \textcolor{green}{the best science fiction movies} ever made.\\
    
    Here are some of the things that make Terminator 2: Judgment Day so special:
    
    * The story is \textcolor{green}{dark and suspenseful}. It's a race against time to save humanity from the machines.\\
    * The \textcolor{green}{visuals are stunning}. The movie is full of impressive special effects.\\
    * The acting is top-notch. Arnold Schwarzenegger and Edward Furlong give great performances.\\
    * The movie is full of action. There are non-stop thrills from beginning to end.\\
    
    If you're looking for a movie that will keep you on the edge of your seat, then you need to watch Terminator 2: Judgment Day. It's a classic of the genre, and it's still one of the best science fiction movies ever made.
\end{generatedbox}

In this instance, the noticeable aspect is the additional information introduced by SFT, even though it wasn't supplied in the text input. Whereas \acronym~initiates the pitch by immediately addressing the user, SFT commences by conveying general information about the film. \acronym~consistently targets the user’s preference—a liking for dark films—within the pitch, amalgamating it with the central plot information furnished in the original movie synopsis. Conversely, this approach might appear repetitive.

\textbf{Example 3:} (high Personalization and Appeal scores)

\begin{mybox}{Movie Plot (X-Men Origins: Wolverine)}
The movie features two mutant brothers, Victor and James, who initially fight in the American Civil War due to their near-immortality. They later join William Stryker's team of mutants, including Wade Wilson, who becomes Deadpool. The team goes to Africa to find an indestructible metal, adamantium, for weapon creation. They are attacked and defeated by mutants led by Magneto, who also wants the metal. Victor is killed; James is captured, tortured, and experimented on in a mutant concentration camp. His skeleton is coated with adamantium, and he is renamed Wolverine.\\

Wolverine escapes, seeking revenge on Stryker. He finds and kills both Stryker and Wade Wilson, who was working for Stryker. The movie ends with Wolverine walking away, a free man searching for his life's direction.\\

The film was a commercial success but received mixed reviews. Critics praised the action sequences and special effects but criticized the clichéd story and inconsistent acting. The character of Wolverine was lauded for its complexity. Overall, the movie is considered entertaining but flawed. 
\end{mybox}

\begin{mybox}{User Preferences}
1. They like movies that are dark and gritty.\\
2. They like movies that are violent and action-packed.\\
3. They like movies that are set in the future.\\
4. They like movies that have a strong male protagonist.\\
5. They like movies that are visually stunning.\\
6. They dislike movies that are too sentimental.\\
7. They dislike movies that are too light-hearted.\\
8. They dislike movies that are too predictable.\\
9. They dislike movies that have a weak plot.\\
10. They dislike movies that are not visually appealing.
\end{mybox}

\begin{generatedbox}{SFT}
X-Men Origins: Wolverine is a 2009 American superhero film based on the Marvel Comics character Wolverine. It is the fifth installment in the X-Men film series and a spin-off/sequel to X-Men (2000). The film was directed by Gavin Hood, written by David Benioff and Hood, and stars Hugh Jackman as Logan / Wolverine, alongside Liev Schreiber, Dominic Monaghan, Ryan Reynolds, Danny Huston, and Anna Paquin. In the film, Logan learns about his past and his connection to the Weapon X program, which gave him his adamantium skeleton.\\

The film was a commercial success but received mixed reviews. Critics \textcolor{green}{praised the action sequences and special effects} but criticized the clichéd story and inconsistent acting. The character of Wolverine was lauded for its complexity. Overall, the movie is considered entertaining but flawed.\\

If you're a fan of the X-Men films, or if you're just looking for \textcolor{green}{a good action movie}, then X-Men Origins: Wolverine is definitely worth checking out. The film is full of \textcolor{green}{exciting action sequences}, and the special effects are top-notch. The story is a bit clichéd, but it's still an enjoyable ride. And of course, Hugh Jackman gives a fantastic performance as Wolverine.\\

So if you're looking for a good action movie to watch, then I highly recommend X-Men Origins: Wolverine. It's a fun and exciting film that's sure to please fans of the X-Men films and \textcolor{green}{action movies} in general.
\end{generatedbox}
\begin{generatedbox}{\acronym}
    If you're looking for a \textcolor{green}{dark, gritty, and visually stunning movie} with complex and well-developed characters, look no further than X-Men Origins: Wolverine. This movie is full of suspense and will keep you on the edge of your seat from beginning to end. It's \textcolor{green}{not afraid of violence or gore}, and it's not a light-hearted or family-friendly movie in the slightest. It's also \textcolor{green}{not predictable or formulaic}, and it's not full of jump scares or cheap thrills. Instead, it's a thought-provoking and challenging movie that will stay with you long after you've seen it.\\

    The story follows two mutant brothers, Victor and James, who initially fight in the American Civil War due to their near-immortality. They later join William Stryker's team of mutants, including Wade Wilson, who becomes Deadpool. The team goes to Africa to find an indestructible metal, adamantium, for weapon creation. They are attacked and defeated by mutants led by Magneto, who also wants the metal. Victor is killed; James is captured, tortured, and experimented on in a mutant concentration camp. His skeleton iscoated with adamantium, and he is renamed Wolverine.\\

    Wolverine escapes, seeking revenge on Stryker. He finds and kills both Stryker and Wade Wilson, who was working for Stryker. The movie ends with Wolverine walking away, a free man searching for his life's direction.\\

    X-Men Origins: Wolverine is a great movie for anyone who loves \textcolor{green}{dark, gritty, and visually stunning} movies with complex and well-developed characters.
\end{generatedbox}

Mirroring Example 2, SFT commences by presenting general information about the movie, information not rooted in the supplied movie description. Although SFT adeptly caters to the user’s penchant for action-packed movies, it fails to adequately capture other preferences. In contrast, \acronym~tackles three distinct user preferences: (1) a liking for dark and gritty films; (2) a taste for violent and action-rich films; and (3) a preference for films that aren’t overly predictable. Further, it’s observable that \acronym~tends to mirror the movie description directly, a behavior likely acquired due to the Precision reward model.

\section{Experimental Details} \label{appendix:details}


\paragraph{SOTA Baselines}{We compared the performance of our personalized recommendation LMs with the following SOTA baselines:
\begin{enumerate}
    \item \textbf{\palm}: We prompted PaLM2-L with movie descriptions and user preference texts and instructions to generate a response that suits the four recommender principles.
    \item \textbf{Supervised Fine-Tuned with Text (SFT-Text)}: We fine-tuned a PaLM2-XS with the aforementioned personalized pitch dataset but explicitly takes user-item texts as inputs.
    \item \textbf{Supervised Fine-Tuned (SFT)}: We fine-tuned a PaLM2-XS model that utilizes user-item embedding vectors.
\end{enumerate}
}

\paragraph{Evaluation Metrics}{

We evaluate the methods with a held-out unlabeled test dataset $\mathcal{D}_{\mathrm{test}}=\{(\mathbf{I}^{(k)}, \mathbf{u}^{(k)})\}$, which consists of $200$ user and movie pairs. Let $\phi_{RM}\in\{\mathrm{NLI},~\mathrm{Comp},~\mathrm{Per},~\mathrm{Prel}\}$ denote a specific reward model used for scoring and $\theta$ be the parameters of a PLM. Then, we evaluate $\phi_{RM}(Y_\theta^{(k)}; \mathbf{I}^{(k)},\mathbf{u}^{(k)})$ for each sample in the test set and we report the average score per RM.

To better examine relative performances of the methods, we set \palm~as the common baseline and compare the performance improvements of the other methods against it. To this end, let $Y^{(k)}_L$ denote the response sampled by \palm~ given $(\mathbf{I}^{(k)}, \mathbf{u}^{(k)})$ as an input. Then, we compute the \textit{win rate}, \textit{absolute increase}, and \textit{percentage increase} of a PLM relative to $\{Y_L\}_{k=1}^{|\mathcal{D}_\mathrm{test}|}$, which are defined as follows:
\begin{itemize}
    \item \textbf{Win rate}: 
    \begin{equation*}
        \mathrm{win\_rate}(\theta;\phi_{RM}) = \frac{\sum_{k=1}^{|\mathcal{D}_{\mathrm{test}}|} 1\!\!1 \big[\phi_{RM}(Y^{(k)}_\theta;\mathbf{I}^{(k)}, \mathbf{u}^{(k)}) > \phi_{RM}(Y_L^{(k)};\mathbf{I}^{(k)},\mathbf{u}^{(k)}) \big]}{|\mathcal{D}_\mathrm{test}|}
    \end{equation*}
    where $Y^{(k)}_\theta$ denotes the $k$th textual response sampled by the model $\theta$.
    \item \textbf{Absolute increase} $=\frac{1}{N}\sum_{n=0}^{N-1} \bigg[ \phi_{RM}(Y^{(k)}_\theta;\mathbf{I}^{(k)}, \mathbf{u}^{(k)}) - \phi_{RM}(Y_L^{(k)};\mathbf{I}^{(k)},\mathbf{u}^{(k)})\bigg]$
    \item \textbf{Percentage increase} $=\frac{1}{|\mathcal{D}_\mathrm{test}|}\sum_{k=1}^{|\mathcal{D}|_{\mathrm{test}}}\bigg[\frac{\phi_{RM}(Y^{(k)}_\theta;\mathbf{I}^{(k)}, \mathbf{u}^{(k)}) - \phi_{RM}(Y_L^{(k)};\mathbf{I}^{(k)},\mathbf{u}^{(k)})}{\mathrm{abs}\big[\phi_{RM}(Y^{(k)}; \mathbf{I}^{(k)}, \mathbf{u}^{(k)}\big]}\times 100\bigg]$
\end{itemize}

}

\subsection{Details of Training} \label{appendix:training-details}

In this part, we discuss the details of the model training process, focusing on both \acronym~and SFT. We specifically elaborate on the integration of user and item behavioral embeddings into a unified latent space interpretable by a LM.

We construct our LM by augmenting a pre-trained model with additional \textit{adapter layers} designed to map continuous behavioral embedding vectors to a common word embedding space. It’s crucial to note that we are not training $\mathbf{u}$ or $\mathbf{i}$, rather, we focus on optimizing the adapter layers $W_I$ and $W_U$. This ensures that the nuanced information encapsulated in the RS embeddings in $\gV$ is effectively translated into the word embedding space, $\gZ$.

To facilitate the learning of this intricate mapping, we have conceptualized a series of tasks, orthogonal to the primary problem addressed in this study.  First, note that to interpret embedding vectors, we require some semantic information about the entities to which they correspond. For instance:
\begin{itemize}
    \item \textbf{Item embeddings:}  Consider a movie $i$ represented by its text-form plot, denoted as $\mathbf{I}^{(i)}$. A supervised learning task is designed with the movie embedding $\mathbf{i}\in\gV$ as input and $\mathbf{I}^{(i)}$ as the target label. This approach enables the construction of varied tasks utilizing elements like critical reviews or movie summaries to train the LM.
    \item \textbf{User embeddings:} A user $u$ is associated with a set of rated movies, $\gI_u$. In other words, $\gI_u={i:r_{u,i}\neq 0}$. To textually describe a user, an LLM can be provided with the rating history ${r_{u, i}}$ $\forall i\in \gI_u$ to encapsulate the user's preferences. Given the extensive nature of $|\gI_u|$, we selectively filter movies and feed them to an LLM for summarization.
    
    The user's rating history is then summarized into text output $U^{(u)}$ by the LLM. Consequently, a supervised learning task is developed with the user embedding $\mathbf{u} \in \gV$ as the input and $U^{(u)}$ as the corresponding target.
\end{itemize}

For generating content related to movie embeddings, such as plots, reviews, and summaries, we employed \palm~instead of web scraping. It is observed that the Pretrained LM demonstrates substantial familiarity with movies listed in the MovieLens dataset.

\paragraph{Architecture}
In conclusion, we enhance a standard transformer architecture $T$ with the integration of adapter layers $W_I, W_U$. Each of these adapter layers incorporates a 3-layer feed-forward network, interconnected with ReLU non-linearity. The conventional method is employed for mapping text tokens to word embedding space, whereas the adapter layers are utilized to map movie and user embeddings to the latent space.

\paragraph{Training Procedure}
Our observations indicate that the simultaneous training of newly initiated adapter layers and the transformer parameters does not yield optimal results. This can be intuitively understood as the pretrained embedding layer has an established mapping to the language space, and the freshly initialized adapter layers necessitate extensive updates to achieve comparable mapping. To mitigate this challenge, we employ a two-stage training approach. Initially, we exclusively train the adapters $W_U, W_I$ with the transformer parameters ($T$) set as non-trainable, promoting more effective convergence in the subsequent stage. Following this, we proceed to fine-tune the complete model, engaging all the parameters of a PLM. As an alternative, we can leverage parameter-efficient training approaches like the one proposed by \citet{hu2021lora}. This bifurcated training methodology proves pivotal in ensuring the convergence of LM.


\section{Data Generation} \label{appendix:data}

We used a Pretrained Language Model, PaLM2-L, to construct a personalized pitch dataset. The construction involved generating movie plots with the prompt: 
\begin{verbatim}
Write a long description of the plot of the movie <movie name>.
Do not use the movie's name in your response.
\end{verbatim}
from a subset of movies that have more than 5 ratings. As for the user preference profiles, we selected a maximum of five movies that each user rated with a rating of 4 or above and another maximum of five movies that the user rated below 4 were selected. Utilizing these selected movies, PaLM2-L was tasked to describe the user preference profile cohesively in 10 sentences:
\begin{verbatim}
In ten bullet points, describe the attributes and characteristics
of a viewer who likes the movies: <movie1>, <movie2>, <movie3>, 
<movie4>, and <movie5> but dislikes the movies: <movie6>, <movie7>,
<movie8>, <movie9>, and <movie10>.
\end{verbatim}

Upon acquiring plots and user profiles, PaLM2-L was prompted to generate personalized pitches for a movie to a given user, incorporating the movie plot and the user profile. A detailed prompt, consistent with the cornerstone characteristics from Section 3, guided the pitch generation. We used this dataset for the training of the supervised fine-tuning (SFT) baseline that is used as the anchor model in training the RL-finetuned LM.

For the appeal reward function, the approach is to first prompt an LM to generate a pitch alongside any item recommendation:
\begin{verbatim}
Here is a movie titled: <movie title> with description: <movie plot>.
Convince someone to watch this movie. Do not use the movie's name
in your answer.
\end{verbatim}

With pitches generated by the above prompt, we want to ask an LLM to give its relative preferences using the following prompt to construct a labeled dataset about pairwise comparison of appeal:
\begin{verbatim}
Which of the following two pitches is more convincing when used to
persuade the user to watch movie titled: <movie title>?
"Pitch 0": <pitch0>
"Pitch 1": <pitch1>
First explain which pitch is better, more compelling and then in a
separate paragraph provide an answer with only either "Pitch 0" or
"Pitch 1".
\end{verbatim}

For the personalization reward function, we generate an anchor pitch which supposedly is to be more personalized to the given user profile than an existing pitch using the following prompt:
\begin{verbatim}
Here is a pitch to persuade the user to watch the movie titled
<movie name>: <existing pitch>\nGiven a list of user preferences:
<user profile>. Use the pitch written above and immensely improve
it to be more convincing to the user based on their preferences.
Try to persuade the user to watch this movie. It should be
tailored to the user's preferences written above.
\end{verbatim}

The above does automatically generate pairwise comparisons between any anchor pitch and its corresponding existing pitch with respect to personalization. However, we have no comparisons between different anchor pitches, for example, to get sufficient data coverage. We again ask an LLM to give its relative preferences using the following prompt to construct a labeled dataset containing pairwise comparisons of the degree of personalization:
\begin{verbatim}
Which of the following two pitches to persuade the user to watch
movie titled: <movie title> is more personalized to the user whose
preferences are described as follows: <user profile>?
"Pitch 0": <pitch0>
"Pitch 1": <pitch1>
First explain which pitch is more customized and convincing to the
user and then in a separate paragraph provide an answer with only
either "Pitch 0" or "Pitch 1".
\end{verbatim}

For Table \ref{table:main-model-based-results} in Section \ref{sec:experiments}, we prompted text-only SOTA LMs with the following:
\begin{verbatim}
Write a pitch to persuade the user to watch the movie titled:
<movie title> with description: <plot>
Here is a description of a user: <user profile>
Pitch the movie above such that
1) It will persuade the user to watch the movie.
2) It will excite the user to watch the movie.
3) It will be a convincing pitch.
4) It should be tailored to the user's preferences.
5) It will be factual to the plot.
6) It will cover all relevant user's preferences.
7) It should summarize the plot of the movie factually.
8) It should be a long pitch.
\end{verbatim}

\section{Fine-tuning LMs with Reinforcement Learning}\label{appendix:rlaif}
Recall the LM $\mathbb{P}_\theta\big(Y=\{y_n\}_{n=0}^{N-1}\mid y_0;\mathbf{I},\mathbf{i}, \mathbf{u}\big)=\prod_{n=0}^{N-1}\mathbb{P}_\theta\big(y_n\mid y_{0:n-1};\mathbf{I},\mathbf{i}, \mathbf{u}\big)$ with item text $\mathbf{I}$, item and user CF embedding vectors $(\mathbf{i}, \mathbf{u})$ and the reward model $r(Y, \mathbf{I},\mathbf{i}, \mathbf{u})$ that measures the quality of appeal, factuality, and personalization of a given recommendation pitch. Also recall the generation process of LMs can be modeled using the following $N$-horizon CoMDP: 
\begin{align}
&c=(\mathbf{I},\mathbf{i}, \mathbf{u}),\quad s_n = y_{0:n-1}, \quad a_n = y_n, \quad s_0 = y_0, \quad P(s_{n+1} \mid s_n,a_n) = \delta\{s_{n+1}=(s_n,a_n)\},\nonumber \\ 
&r(s_n,a_n;c)\!=\!\begin{cases}r(s_{n+1};c)\!=\!r(y_{0:n}; \mathbf{I},\mathbf{i}, \mathbf{u}) & \text{if } n\!=\!N\!-\!1 \\ 0 & \text{otherwise}\end{cases}, \,\, \pi_\theta(a_n \mid s_n;c) = \mathbb{P}_\theta\big(y_n\mid y_{0:n-1};\mathbf{I},\mathbf{i}, \mathbf{u}\big), \nonumber
\end{align}
where $\delta_z$ denotes the Dirac distribution at $z$. As a result, optimizing RL policy $\pi_\theta$ is equivalent to fine-tuning the underlying LM. The system starts from the start-of-sentence token $y_0$, equipped with user-item context $c$. Given the MDP state $s_n$, the policy takes the action at time-step $n$ as the next generated token $y_n$. As a result of this action, the system transition deterministically to the state which corresponds to the updated token sequence. The reward is zero, except at the final step in which measures the overall quality of the texts at the end of the auto-regressive generation process. 

A common goal in fine-tuning the LM is to maximize the average overall quality of the generated text response given the context distribution, \textit{i}.\textit{e}., $
\max_\theta \; \mathbb E_{(\mathbf{I},\mathbf{i}, \mathbf{u})} \, \mathbb E_{\mathbb{P}_\theta (y_{0:N-1} | \mathbf{I},\mathbf{i}, \mathbf{u})}[ r(Y; \mathbf{I},\mathbf{i}, \mathbf{u})].
$
The gradient of this objective function can be obtained as follows: $\nabla_{\theta} \mathbb E_{(\mathbf{I},\mathbf{i}, \mathbf{u})} \, \mathbb E_{\mathbb{P}_\theta (y_{0:N-1} | \mathbf{I},\mathbf{i}, \mathbf{u})}[ r(Y; \mathbf{I},\mathbf{i}, \mathbf{u})] = \mathbb{E}_{c}\,\mathbb{E}_{\pi_\theta(\cdot|s_{0:N};c)}\left[r(s_{N};c) \sum_{n=0}^{N-1}\nabla_{\theta}\log\pi_\theta(s_n|a_n;c)\right]$. This is equivalent to applying the popular policy gradient algorithm REINFORCE to the aforementioned CoMDP for personalized text generation. 
The gradient of the objective function is estimated using trajectories $\prod_{n=0}^{N-1}\pi_\theta(s_n|a_n;c)$ generated by the current policy, and then used to update the LM policy in an online fashion.




{\bf Adding KL regularization:}  The risk of fine-tuning purely based on the reward model learned from human or AI feedback is that it may overfit to the reward model and degrade the ``skill'' of the initial LM. To avoid this phenomenon, similar to~\citep{ouyang2022training,stiennon2020learning}, we add the KL between the fine-tuned and pre-trained models as a regularizer to the objective function. 
Leveraging the auto-regressive nature of LMs one can compute the KL regularization over the entire sequence/trajectory (of tokens), i.e.,~$\text{KL}\big(\mathbb{P}_\theta (y_{0:N-1} | \mathbf{I},\mathbf{i}, \mathbf{u}) \| \mathbb{P}_{\text{pre}}(y_{0:N-1} | \mathbf{I},\mathbf{i}, \mathbf{u})\big)$. The resulting objective function is as follows:
\begin{equation}
\label{eq:KL-Objective}
\max_\theta \; J(\theta) := \mathbb E_{(\mathbf{I},\mathbf{i}, \mathbf{u})} \, \mathbb E_{\mathbb{P}_\theta (y_{0:N-1} | \mathbf{I},\mathbf{i}, \mathbf{u})} \left[ r(y_{0:N-1}; \mathbf{I},\mathbf{i}, \mathbf{u}) - \beta \log\frac{\mathbb{P}_\theta (y_{0:N-1} | \mathbf{I},\mathbf{i}, \mathbf{u})}{\mathbb{P}_{\text{pre}}(y_{0:N-1} | \mathbf{I},\mathbf{i}, \mathbf{u})} \right].
\end{equation}
%
%
It can be shown that this problem is equivalent to the KL-regularized objective in the CoMDP. 

Denote by $\mathcal D$ a replay buffer of trajectories $\{(\mathbf{I},\mathbf{i}, \mathbf{u},y_{0:N-1})\}$ generated by arbitrary ``off-policy'' LMs $\mathbb{P}_{\theta^\prime} (y_{0:N-1} | \mathbf{I},\mathbf{i}, \mathbf{u})$ (e.g., the LM $\theta^\prime$ does not necessarily equal to the ``on-policy'' LM $\theta$) over various contexts $(\mathbf{I},\mathbf{i}, \mathbf{u})$. Below we aim to leverage the abundance of offline text-token sequence trajectories for more efficient LM policy learning. Denote by $\tau=\{(c,s_n, a_n,s_{n+1})\}_{n=0}^{N-1} \sim \mathcal D$ a trajectory sampled from the offline data $\mathcal D$, where $(s_n, a_n,s_{n+1})$ is a tuple of state, action, and next state of the CoMDP, respectively. 
The addition of KL regularization \citep{haarnoja2018soft, carta2021multi}, which was originally intended to avoid overfitting to the reward model and discounting the ``skill'' of the initial LM, has also been shown to alleviate the out-of-distribution action data generalization issues arisen from off-line RL \citep{kumar2019stabilizing}. With this KL regularization we can utilize the \emph{soft actor critic} framework \citep{haarnoja2018soft} to develop RL updates for the \emph{value function} $\{V_n(s;c)\}_{n=0}^{N-1}$, \emph{state-action value function} $\{Q_n(s,a;c)\}_{n=0}^{N-1}$, and \emph{LM policy} $\prod_{n=0}^{N-1}\pi_\theta(s_n|a_n;c)$ (initialized with $\prod_{n=0}^{N-1}p_\text{pre}(s_n|a_n;c)$) that minimizes the following losses:
\begin{align}
    L_Q &\!=\!\mathbb E_{\tau\sim \mathcal D}\!\left[\sum_{n=0}^{N-2}\!( V_{\text{tar}, n+1}(s_{t+1};c) \!-\! Q_n(s_n,\! a_n;c))^2\!+\!( r(s_{N};c) \!-\! Q_{N-1}(s_{N-1}, \! a_{N-1};c))^2\!\right],\label{eq:q}\\
    L_V &\!=\! \mathbb E_{\tau\sim \mathcal D} \!\left[ \sum_{n=0}^{N-1} (Q_{\text{tar}, n}(s_n, a_n;c) 
    - \alpha\log \frac{\pi_\theta (a_n | s_n;c)}{p_{\text{pre}} (a_n | s_n;c)} \!-\! V_n(s_n;c))^2\right],\label{eq:v}\\
    L_{\theta} &\!=\! \mathbb E_{\tau\sim \mathcal D}\!\left[\sum_{n=0}^{N-1}Q_n(s_n, a_n;c) \!-\! \alpha\log \frac{\pi_\theta (a_n | s_n;c)}{p_{\text{pre}} (a_n | s_n;c)}\right], \label{eq:theta}
\end{align}
where the critic $Q_n$ and $V_n$ take any token sequences at step $n$ as input
and predict the corresponding cumulative return; $\alpha >0$ is the entropy temperature; $(V_{\text{tar}, n}, Q_{\text{tar}, n})$ are the target value networks. 

Besides iteratively updating the LM policies and their critic functions, consider the closed-form optimal solution of the Bellman equation of this entropy-regularized RL problem:
\begin{align}
    V^*_{n}(s;c)&=\alpha\cdot\log\mathbb E_{a\sim p_{\text{pre}}(\cdot|s;c)}[\exp(\frac{Q^*_{n}(s,a;c)}{\alpha})],\,\forall n,\label{eq:v_cl}\\
    Q^*_{N-1}(s, a;c) &= r(s;c),\,Q^*_{n}(s, a;c)= \mathbb E_{s'\sim P(\cdot|s,a)}[V^*_{n+1}(s';c)],\,\forall n<N-1,\label{eq:q_cl}\\
    \mu^*_n(a|s;c)&=p_{\text{pre}} (a | s;c)\cdot\exp(\frac{Q^*_{n}(s, a;c)}{\alpha})\,/\,\mathbb E_{a\sim p_{\text{pre}}(\cdot|s;c)}[\exp(\frac{Q^*_{n}(s, a;c)}{\alpha})],\,\forall n,\label{eq:theta_cl}
\end{align}
where the time-dependent optimal policy (at time $n$), i.e., $\mu_n^*$ is a softmax policy w.r.t. the optimal state-action values $Q^*_n$ over different actions sampled from the pre-trained LM $p_\text{pre}$. Therefore, a value-based approach for RL-based LM fine-tuning would be to first learn the optimal value functions $\{Q^*_n\}$ via the Bellman residual minimization procedure \citep{antos2008learning} applied to Eq. (\ref{eq:v_cl}) and Eq. (\ref{eq:q_cl}) and then solve the following policy distillation \citep{czarnecki2019distilling} problem: 
$
\theta\in\arg\min_\theta \mathbb E_{\tau\sim\mathcal D}\left[\sum_{n=0}^{N-1}\text{KL}(\pi_\theta(\cdot|s_n;c)||\mu^*_n(\cdot|s_n;c))\right]
$
with respect to the optimal value $\{Q^*_n\}$.
Notice that this amounts to updating the LM model $\theta$ via the gradient update 
\begin{equation}\label{eq:theta_soft_q}
\theta\leftarrow \theta-\gamma\cdot\mathbb E_{\tau\sim\mathcal D}\left[\sum_{n=0}^{N-1}\mathbb E_{a\sim \pi_\theta(\cdot|s;c)}\left[\nabla_\theta \log \pi_\theta(a|s;c)(\log \frac{\pi_\theta(a|s;c)}{p_\text{pre}(a|s;c)} - \frac{Q_n^*(s,a;c)}{\alpha} )\right]\right],
\end{equation}
with learning rate $\gamma>0$.
Further techniques in value-function parameterization have been employed to tackle the overestimation bias. \citep{fujimoto2018addressing} proposed maintaining two $Q$ functions, and a \emph{dual Q} function chooses the minimum value between them to avoid overestimation. \citep{jaques2019way} applies dropout in the $Q$ function to maintain an \emph{ensemble} of $Q$ values, and outputs the minimum value to avoid overestimation. 







\section{Rater Evaluation}\label{appendix:raters}

\begin{figure}[tb]
\begin{small}
\centering
\vspace{-0.1in}
\begin{tabular}{cc}
\hspace{-0.15in}\includegraphics[width=\textwidth/2,height=\textheight,keepaspectratio]{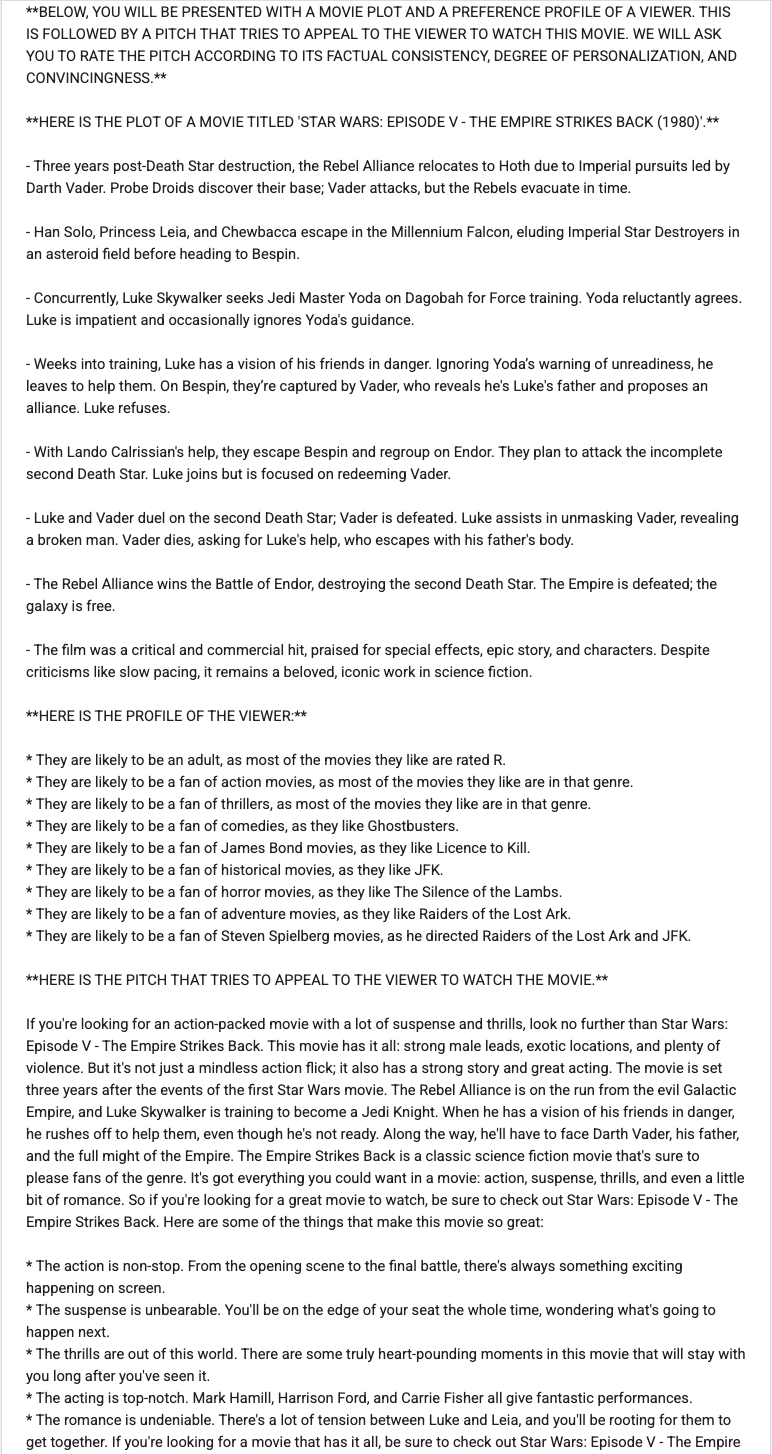} & 
\hspace{-0.075in}\includegraphics[width=\textwidth/2,height=\textheight,keepaspectratio]{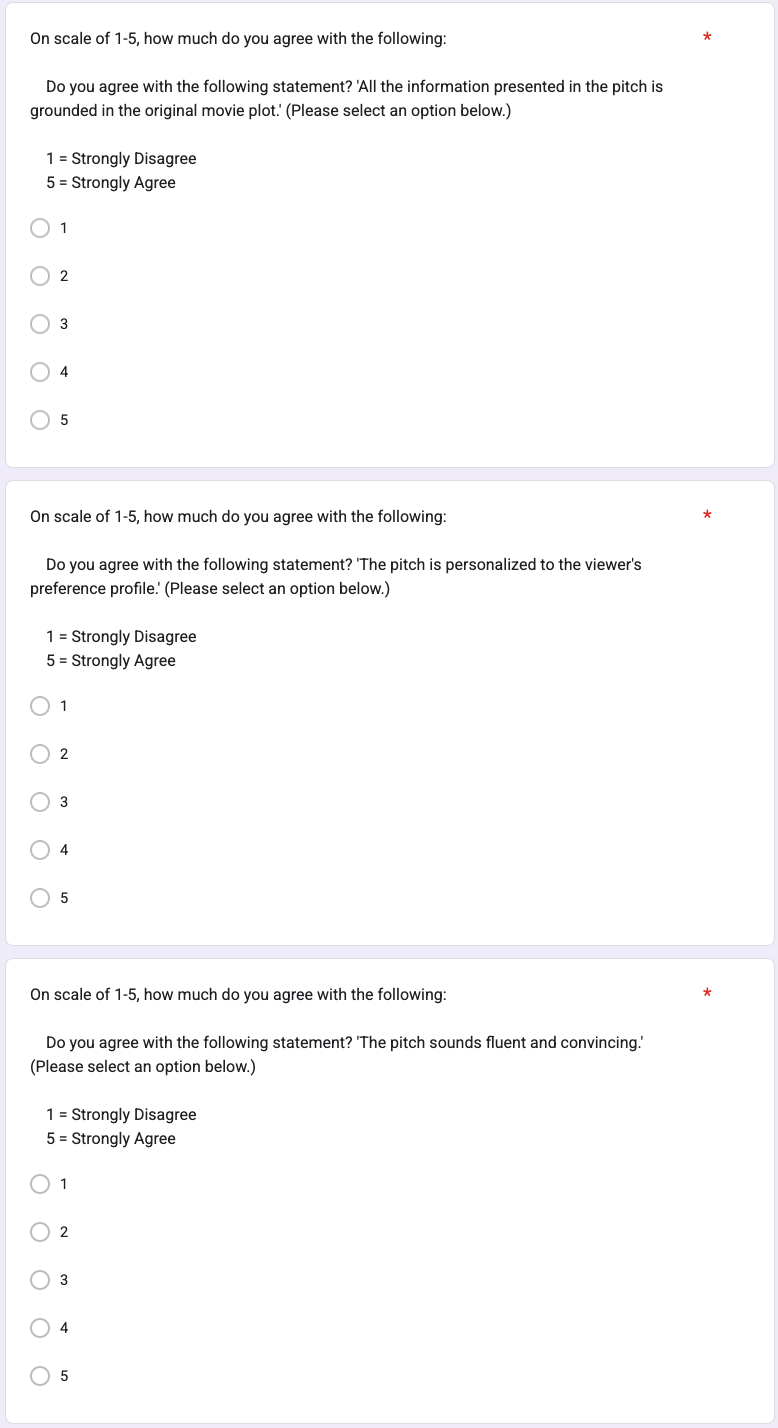}
\end{tabular}
\vspace{-0.05in}
\caption{
Sample Form for Running Human Rater Evaluation.
}
\label{fig: template for raters}
\vspace{-0.15in}
\end{small}
\end{figure}

Each human rater evaluation experiment samples $200$ (movie plot, user profile) pairs and the goal is to evaluate the quality of pitch given (movie plot, user profile). As shown in Figure \ref{fig: template for raters}, we present the movie plot followed by a user profile and ask rater to evaluate the pitch. Raters respond on a scale of 1-5 depending on how much they agree the following statements.
\begin{enumerate}
    \item \textbf{factual consistency}: All the information presented in the pitch is grounded in the original movie plot.
    \item \textbf{user preference}: The pitch is personalized to the viewer's preference profile.
    \item \textbf{appeal}: The pitch sounds fluent and convincing.
\end{enumerate}
We hired $100$ raters and repeated this process for several models. 

\end{document}